\documentclass[10pt]{article} 
\usepackage[preprint]{rlc}

\usepackage{amssymb}            
\usepackage{mathtools}          
\usepackage{mathrsfs}           
\mathtoolsset{showonlyrefs}     
\usepackage{graphicx}           
\usepackage{subcaption}         
\usepackage[space]{grffile}     
\usepackage{url}                
\usepackage{amsthm}
\usepackage{todonotes}
\presetkeys{todonotes}{inline,backgroundcolor=yellow!40,caption={}}{}
\usepackage{cleveref}
\usepackage{multirow}
\usepackage{booktabs} 
\usepackage{siunitx} 
\usepackage{tabularx} 
\usepackage{amsmath}
\usetikzlibrary{positioning,arrows.meta,fit,calc,backgrounds,decorations.pathmorphing}
\usepackage{paralist}

\renewcommand*\d{\mathop{}\!\textnormal{\slshape d}}

\newcommand{\describe}[2]{\underbracket[1pt][0pt]{#1}_\text{\makebox[1em][c]{#2}}}

\DeclarePairedDelimiterX{\SquareBrackets}[1]{[}{]}{#1}
\DeclarePairedDelimiterX{\RoundBrackets}[1]{(}{)}{#1}
\DeclarePairedDelimiterX{\DivergenceBrackets}[2]{[}{]}{#1\;\delimsize\|\;#2}

\NewDocumentCommand{\pr}{ O{p} r() }{
  \def\prArg{#2}\patchcmd{\prArg}{|}{\mid}{}{}#1\RoundBrackets{\prArg}}
\NewDocumentCommand{\p}{ r() }{\pr[p](#1)}
\NewDocumentCommand{\q}{ r() }{\pr[q](#1)}
\NewDocumentCommand{\prm}{ r() }{\pr[\mathrm{p}](#1)}
\NewDocumentCommand{\Normal}{ r() }{\pr[\operatorname{Normal}](#1)}
\NewDocumentCommand{\Cat}{ r() }{\pr[\operatorname{Cat}](#1)}
\NewDocumentCommand{\Beta}{ r() }{\pr[\operatorname{Beta}](#1)}
\NewDocumentCommand{\Bernoulli}{ r() }{\pr[\operatorname{Bernoulli}](#1)}
\NewDocumentCommand{\Dir}{ r() }{\pr[\operatorname{Dir}](#1)}

\newcommand{\E}[3][]{\mathbb{\operatorname{E}}_{#2}#1[#3#1]}

\newcommand{\KL}{\mathrm{\operatorname{KL}}\DivergenceBrackets}
\newlength\widthE
\newcommand{\Ebelow}[3][]{\settowidth\widthE{$\operatorname{E}$}
\mathop{\operatorname{E}}_{\vphantom{|^|}\mathmakebox[0.5\widthE][l]{#2}}#1[#3#1]}

\definecolor{default}{rgb}{0.3137254901960784,0.4627450980392157,0.6588235294117647}
\definecolor{hidden}{rgb}{0.8,0.5372549019607843,0.38823529411764707}
\definecolor{concat}{rgb}{0.2784313725490196,0.6274509803921569,0.4117647058823529}
\definecolor{crssm}{rgb}{0.7607843137254902,0.3411764705882353,0.3686274509803922}
\definecolor{ep}{rgb}{0.16470588235294117,0.6666666666666666,0.3803921568627451}
\definecolor{rp}{rgb}{0.39215686274509803,0.7098039215686275,0.803921568627451}
\definecolor{fg}{rgb}{0.10,0.6098039215686275,0.25}

\usepackage{xspace}
\newcommand{\crssm}{cRSSM\xspace}
\newcommand{\defaultagent}{default-context\xspace}
\newcommand{\hiddenagent}{hidden-context\xspace}
\newcommand{\concatagent}{concat-context\xspace}
\newcommand{\ccrssm}{{\color{crssm}\crssm}}
\newcommand{\cdefaultagent}{{\color{default}\defaultagent}}
\newcommand{\chiddenagent}{{\color{hidden}\hiddenagent}}
\newcommand{\cconcatagent}{{\color{concat}\concatagent}}
\newcommand{\expertpol}{{\color{ep}expert}\xspace}
\newcommand{\randompol}{{\color{rp}random}\xspace}

\title{Dreaming of Many Worlds: Learning Contextual World Models Aids Zero-Shot Generalization}

\author{Sai Prasanna\thanks{equal contribution}\hspace{14.9mm} Karim Farid{\color{dark-blue}\footnotemark[1]}\hspace{14.9mm} Raghu Rajan\hspace{14.9mm} André Biedenkapp\\
\{ramans, faridk, rajanr, biedenka\}@cs.uni-freiburg.de\\
University of Freiburg\vspace{-10mm}
}

\begin{document}

\maketitle

\begin{abstract}

Zero-shot generalization (ZSG) to unseen dynamics is a major challenge for creating generally capable embodied agents. To address the broader challenge, we start with the simpler setting of contextual reinforcement learning (cRL), assuming observability of the context values that parameterize the variation in the system's dynamics, such as the mass or dimensions of a robot, without making further simplifying assumptions about the observability of the Markovian state. Toward the goal of ZSG to unseen variation in context, we propose the contextual recurrent state-space model (cRSSM), which introduces changes to the world model of Dreamer~(v3) \citep{hafner-arxiv23a}. This allows the world model to incorporate context for inferring latent Markovian states from the observations and modeling the latent dynamics. Our approach is evaluated on two tasks from the CARL benchmark suite, which is tailored to study contextual RL. Our experiments show that such systematic incorporation of the context improves the ZSG of the policies trained on the ``dreams'' of the world model. We further find qualitatively that our approach allows Dreamer to disentangle the latent state from context, allowing it to extrapolate its dreams to the many worlds of unseen contexts. The code for all our experiments is available at \url{https://github.com/sai-prasanna/dreaming_of_many_worlds}.
\end{abstract}

\section{Introduction}
Model-Based Reinforcement Learning (MBRL) promises to be one of the most data-efficient frameworks for learning control.
With this data efficiency, MBRL could significantly impact real-world applications, such as robotics and autonomous systems, for which efficient learning and generalization are paramount.
Recent MBRL approaches are capable of achieving performance comparable to model-free reinforcement learning (MFRL) algorithms while only requiring a fraction of the data \citep[see, e.g.,][]{chua-neurips18a,hafner-iclr20a,hafner-iclr2021,hafner-arxiv23a,wu-corl22a,hansen-iclr24a}.

A key challenge in MBRL is the ability to generalize to unseen environments, particularly in a \textit{zero-shot} setting, where an agent must perform effectively in novel scenarios without prior experience.
Although MBRL has shown great improvement in recent years, both MBRL and MFRL algorithms remain susceptible to small changes in environment dynamics \citep{kirk-jair23a}.
This can be attributed in part to the complexity of the (MB)RL pipeline \citep{zhang-aistats21a,parkerholder-jair22a} but also to a lack of understanding, as zero-shot generalization (ZSG) remains an understudied domain for MBRL \citep{kirk-jair23a}.

An influential family of MBRL algorithms is Dreamer \citep{hafner-iclr20a, hafner-iclr2021, hafner-arxiv23a}.
Dreamer-like algorithms learn a latent representation of the world from which plausible trajectories can be imagined that can be used to improve decision-making. The family of Dreamer algorithms has achieved impressive results in various domains in both learned policy performance and sample efficiency during learning.
However, Dreamer-like algorithms have not yet been studied in the zero-shot generalization setting.

Here, we propose to use the contextual reinforcement learning paradigm \citep{kirk-jair23a,benjamins-tmlr23a} to study Dreamers' learning capabilities within and across many worlds.
To this end, we assume that we have access to privileged information about how the transition dynamics of the underlying Markov decision process (MDP) is parameterized, i.e., the context\citep{hallak-arxiv15a}.
We use tasks from the Contextual and Adaptive Reinforcement Learning benchmark \citep[CARL;][]{benjamins-tmlr23a} where the context defines some physical properties that affect the dynamics and that an RL agent can observe while trying to solve a given task. Examples of such properties are gravity or the mass of a load that a robot might lift. We assume the context parameters are continuous and hence can be studied meaningfully for interpolation and extrapolation, unlike discrete parameters where generalization to unseen values is not well-defined.

We analyze Dreamer's ZSG capabilities, in- and out-of-distribution (OOD), when naively integrating context, and we propose an improved Dreamer variant that integrates context more intelligently and demonstrates improved generalization abilities.
In particular, our contributions are as follows.
\begin{itemize}\itemsep0pt
    \item We provide the first principled study in understanding Dreamer's generalization capabilities for in- as well as out-of-distribution (OOD) tasks on two tasks from CARL;
    \item We propose a novel approach for conditioning the Dreamer architecture on context and show how it improves Dreamer's zero-shot generalization ability on the given tasks; and
    \item We show in a case study how our approach to context-conditioning shapes and improves Dreamer's imagination capabilities.
\end{itemize}

\section{Related Work}

Our approach aims to improve the ZSG of MBRL agents. As such, in this section, we discuss related works from meta-RL, an area aimed at improving few- and zero-shot generalization, followed by MBRL and ZSG in MBRL.

\paragraph{Meta-RL} Meta-reinforcement learning (meta-RL) \citep{beck2023survey} has been proposed as a promising approach to address the challenge of generalization in RL. Meta-RL aims to learn an RL agent that can adapt to new tasks in a sample-efficient manner. Meta-RL algorithms \citep[see, e.g.,][]{duan-arxiv16a,wang-cogsci17a,nagabandi2018learning,rakelly-icml19a,melo-icml22a,wen-arxiv23a} are designed to quickly adapt to new and unseen settings with limited access to new experiences (i.e., few-shot adaptation) generated by the RL agent. In contrast, our work focuses on zero-shot generalization (ZSG) for RL \citep{kirk-jair23a}, where we aim to learn policies that are capable of zero-shot adaptation to new settings without assuming access to further training or the reward signal. 

\paragraph{Model-Based RL}
MBRL is believed to be one of the most promising directions to improve the sample efficiency of RL algorithms. \citet{young2022benefits} make the case that algorithms that use experience with a model can generalize to unseen environments better than those that rely purely on value-function generalization and experience replay. 

Empirically, MBRL algorithms, such as Dreamer \citep{hafner-iclr20a,hafner-arxiv23a} and TD-MPC2 \citep{hansen-iclr24a} pipelines, achieve state-of-the-art sample efficiency. While Dreamer's focus was on achieving a high return in a variety of individual environments, our focus is to study and improve generalization capabilities in contextual variants of environments and we evaluate our work on currently available environments tailored for this. Based on Dreamer's success, we build our approach on it and study its zero-shot generalization capabilities before suggesting an improved approach. Dreamer has recently been studied in the meta-RL case for a few-shot generalization \citep{wen-arxiv23a}. However, this work still requires many interactions with the target domain for the agent to learn to adapt to a test task successfully.

\paragraph{Zero-Shot Generalization in MBRL}
Studies on zero-shot generalization (ZSG) have mainly focused on the model-free case \citep{kirk-jair23a}. The few works that have studied zero-shot generalization in MBRL assume that the context is not observable by an agent and would need to be inferred for agents to adapt \citep{lee2020context, perez2020generalized, zhang-iclr21a, ball-icml21a, guo2022a, sodhani2022block, wen-arxiv23a}.
In contrast, our work is more similar to the study of \citet{benjamins-tmlr23a}, which assumes that the context is observable and accessible by an agent.
\citet{benjamins-tmlr23a} evaluated multiple model-free agents for ZSG. \citet{beukman-neurips23a} build on this approach and propose to learn a hypernetwork to adapt a SAC~\citep{haarnoja-icml18a} agent policy based on the observed context.
We believe that the observable context setting holds a lot of merit, as, on the one hand, various physical contexts that can be sensed by a real robot, such as the mass of a load it carries, could be used to improve its policy in different contexts \citep{escontrela-arxiv20a}. On the other hand, insights gained in this observable context setting will likely be useful for the more challenging setting where context is hidden and needs to be inferred.

Furthermore, while our work assumes an observable context, it still tackles the challenging setting of partial observability in the underlying latent Markovian state. Our work contrasts previous work in ZSG that operates under the assumption that the latent state is observable or can be decoded from purely high-dimensional observations \citep{du2019provably}. To tackle such partial observability, we design a systematic method to use context to estimate latent states when the context is visible.
An early work in this direction learns policies for helicopter control under partial observability \citep{rogier2009}. They estimate the latent state (wind) from the known context parameters thereby improving their policy performance.

\section{Background}\label{sec:back}
\paragraph{Zero-Shot Generalization in Contextual Markov Decision Processes}
To empirically study ZSG in a partially observable setting, we use a definition of the contextual MDP~\citep[cMDP;][]{hallak-arxiv15a,modi-alt18a} similar to the one proposed by \citet{kirk-jair23a}.
A cMDP is a tuple $
    M := (S, A, O, R, T, C, \phi, \p(s_0|c), \p(c))
$ 
where $S$ is the state space, $A$ is the action space, $O$ is the observation space,  $R\colon S \times C \times A \to \mathbb{R}$ is the reward function, $T\colon S \times A \times C \to \text{dist} (S)$ is the stochastic Markov transition function over the states. $C$ is the context space. 
$\phi\colon S \times C \to O$ is the observation emission function, $\p(s_0|c)$ is the initial state distribution for a given context, and $\p(c)$ is the context distribution with $c\in C$. For the commonly used discrete-time case, the timesteps are $t \in [0, H]$, with $H$ as the horizon per episode. Context remains the same during an episode, but may change across episodes.

For a given cMDP, we can train a policy $\pi: \p(a_t |o_{\le t},a_{<t}, c)$ that is trained with the objective of maximizing the expected sum of rewards $\mathbb{E}_\pi(\sum_{t=0}^{H} r_t)$ in the distribution of the training contexts $p_{\text{train}}(c)$. We can then study the generalizability of this policy in the zero-shot setting by evaluating transfer on an evaluation context distribution $p_{\text{eval}}(c)$.

\paragraph{Importance of Context}
We often face partial observability of latent Markovian states in real-world RL tasks. Providing context to an agent may help infer such latent Markovian states. Generally, context refers to an aspect of the environment or the agent that remains constant for a certain period and affects its behavior. In our work, we focus on context that remains unchanged throughout the entire episode.

As a motivating example, consider a wheeled robot that has to deliver goods to various locations, encountering varying terrains along the way, including rough terrains that may damage the robot's wheels and, ultimately, the robot. Now, assume that the robot has a sensor to measure its velocity. If the robot must navigate without damaging itself, it may be useful to estimate the coefficient of friction of the surface on which it moves. This coefficient of friction is a non-stationary latent state variable that changes within an episode. Here, the mass of the load on the robot (context), the coefficient of friction of the terrain (latent state), and the torque applied to the wheels (action) causally affect the velocity (observation) of the robot. The robot must use its context, observations, and actions to infer the friction. The robot could then use this estimate of the friction to improve its policy and decide whether to apply more or less torque.  
This example, though oversimplified, gives us an idea of how effective the context might be to infer latent states and improve the policy in partially observable settings. 
It motivates the design of our contextual Dreamer agent.

\section{Method}

We now discuss how we incorporate context into the Dreamer~(v3) \citep{hafner-arxiv23a} algorithm. We first introduce our novel approach to contextual dreaming and then contrast it with naive ways of learning from and with context information.

\subsection{Contextual Dreamer}
\label{subsec:Dreamer}
We employ a novel contextual recurrent state space model (cRSSM) that builds on Dreamer's RSSM world model and systematically introduces context. Here, we discuss how it can be used to imagine trajectories in the contextual RL setting and describe how we alter Dreamer's actor-critic policy network to use the context and the latent states inferred by the contextual world model.

\subsubsection{Contextual Recurrent State-Space Model (cRSSM)}
\begin{figure*}[ht]\vskip-1.66cm
\centering
\hfil\hfil
\begin{subfigure}[t]{.49\textwidth}
\centering
\scalebox{0.66}{%
\begin{tikzpicture}[
  node distance=2.5em, auto,
  lat/.style={draw=black, circle, minimum size=2em},
  det/.style={draw=black, rectangle, minimum size=2em},
  obs/.style={circle, draw=black, fill=black!20, minimum size=2em},
  gen/.style={->, -{Stealth[length=.5em, inset=0pt]}},
  ctx/.style={circle, draw=crssm, fill=black!20, minimum size=2em},
  gen_ctx/.style={->, draw=crssm, -{Stealth[length=.5em, inset=0pt]}},
]
\node[obs, inner sep=.02em] (o1) {$o_1,r_1$};
\node[obs, right=of o1, inner sep=.02em] (o2) {$o_2,r_2$};
\node[lat, right=of o2, inner sep=.02em] (o3) {$o_3,r_3$};
\node[lat, above=of o1] (s1) {$s_1$};
\node[lat, above=of o2] (s2) {$s_2$};
\node[lat, above=of o3] (s3) {$s_3$};
\node[obs, above=of s1] (a1) {$a_1$};
\node[lat, above=of s2] (a2) {$a_2$};
\path (s1) edge[gen] node {} (o1);
\path (s2) edge[gen] node {} (o2);
\path (s3) edge[gen] node {} (o3);
\path (s1) edge[gen] node {} (s2);
\path (s2) edge[gen] node {} (s3);
\path (a1) edge[gen] node {} (s2);
\path (a2) edge[gen] node {} (s3);
\node[ctx, left=of a1] (c) {$c$};

\path (c) edge[gen_ctx] (s1);
\path (c) edge[gen_ctx] (s2);
\path (c) edge[gen_ctx] (s3);
\path (c) edge[gen_ctx] (o1);
\path (c) edge[gen_ctx, bend left=19] (o2);
\path (c) edge[gen_ctx, bend left=90, looseness=1.25] (o3);
\end{tikzpicture}}
\caption{Generative Model for a cMDP}
\label{fig:cMDP-gen}
\end{subfigure} \hfil%
\begin{subfigure}[t]{.49\textwidth}
\centering
\scalebox{0.66}{%
\begin{tikzpicture}[
  node distance=2.5em, auto,
  lat/.style={draw=black, circle, minimum size=2em},
  det/.style={draw=black, rectangle, minimum size=2em},
  obs/.style={circle, draw=black, fill=black!20, minimum size=2em},
  ctx/.style={circle, draw=crssm, fill=black!20, minimum size=2em},
  gen/.style={->, -{Stealth[length=.5em, inset=0pt]}},
  gen_ctx/.style={->, draw=crssm, -{Stealth[length=.5em, inset=0pt]}},
  inf/.style={dashed, ->, -{Stealth[length=.5em, inset=0pt]}},
]

\node[obs, inner sep=.02em] (o1) {$o_1,r_1$};
\node[obs, right=of o1, inner sep=.02em] (o2) {$o_2,r_2$};
\node[lat, right=of o2, inner sep=.02em] (o3) {$o_3,r_3$};
\node[lat, above=of o1] (z1) {$z_1$};
\node[lat, above=of o2] (z2) {$z_2$};
\node[lat, above=of o3] (z3) {$z_3$};
\node[det, above=of z1] (b1) {$h_1$};
\node[det, above=of z2] (b2) {$h_2$};
\node[det, above=of z3] (b3) {$h_3$};
\node[obs, above=of b1] (a1) {$a_1$};
\node[lat, above=of b2] (a2) {$a_2$};
\path (z1) edge[gen] node {} (o1);
\path (z2) edge[gen] node {} (o2);
\path (z3) edge[gen] node {} (o3);
\path (b1) edge[gen, bend left=35] node {} (o1);
\path (b2) edge[gen, bend left=35] node {} (o2);
\path (b3) edge[gen, bend left=35] node {} (o3);
\path (b1) edge[gen] node {} (z1);
\path (b2) edge[gen] node {} (z2);
\path (b3) edge[gen] node {} (z3);
\path (b1) edge[gen] node {} (b2);
\path (b2) edge[gen] node {} (b3);
\path (z1) edge[gen] node {} (b2);
\path (z2) edge[gen] node {} (b3);
\path (a1) edge[gen] node {} (b2);
\path (a2) edge[gen] node {} (b3);
\node[ctx, left=of a1] (c) {$c$};

\path (c) edge[gen_ctx] (b1);
\path (c) edge[gen_ctx] (b2);
\path (c) edge[gen_ctx] (b3);
\path (c) edge[gen_ctx] (o1);
\path (c) edge[gen_ctx, bend right=3] (o2);
\path (c) edge[gen_ctx, bend left=95, looseness=1.4] (o3);

\path (b2) edge[inf, bend right=30] node {} (z2);
\path (o2) edge[inf, bend left=30] node {} (z2);
\end{tikzpicture}}
\caption{cRSSM}
\label{fig:cRssm}
\end{subfigure}%
\hfil\hfil %
\caption{Latent dynamics models.
The models shown observe the first two time steps and predict the third.
Circles represent stochastic variables, and squares represent deterministic variables. Solid lines denote the generative process, and dashed lines denote the inference model. The context node and edges are highlighted in {\color{crssm}red}. 
(\protect\subref{fig:cMDP-gen}) The generative model for a cMDP. 
(\protect\subref{fig:cRssm}) Our cRSSM.
}
\label{fig:model}
\end{figure*}
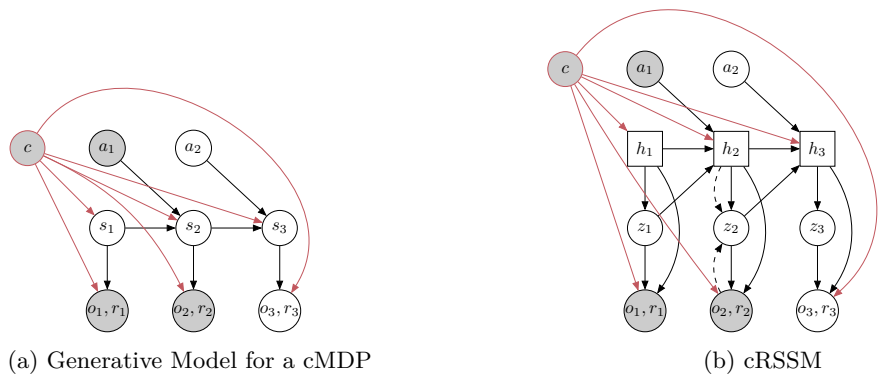

We first define a non-linear latent space model (see \Cref{fig:cMDP-gen}) for the general formulation of a cMDP (see \Cref{sec:back}) with partial observability. This defines the generative process of observations ${\{o_t\}}_{t=1}^{H}$ and rewards ${\{r_t\}}_{t=1}^{H}$ from latent states ${\{s_t\}}_{t=1}^{H}$, actions ${\{a_t\}}_{t=1}^{H}$, and context $c$. This generative model describes the influence of context on the transition dynamics, rewards, and observations. 

\paragraph{World Model Objective} To perform inference of the latent states for this non-linear model, we cannot directly compute the posterior \citep{hafner2019learning}. Instead, we learn an encoder $\q(s_{1:H}|o_{1:H},a_{1:H}, {\color{red}c} )=\prod_{t=1}^H \pr[q_\theta](s_t|s_{t-1},a_{t-1},o_t, {\color{red}c})$. This encoder incorporates context to estimate latent states from observations and actions. Using the encoder, we follow \citet{hafner2019learning} in constructing a variational bound on the data log-likelihood. Here, we write the objective for predicting only the observations; a similar derivation applies for the rewards and the prediction of the continuation flag\footnote{Continuation flag indicates whether the state is terminal.} of the episode $n_t \in {0, 1}$. The evidence lower bound (ELBO) obtained using Jensen's inequality is then
{\small\begin{multline}
\label{eq:cRSSM-elbo}
\ln\p(o_{1:T}|a_{1:T}, {\color{red}c})
\triangleq\ln\int\prod_t\p(s_t|s_{t-1},a_{t-1}, {\color{red}c})\p(o_t|s_t, {\color{red}c})\d s_{1:T} \\[-2ex]
\begin{aligned}
\geq\sum_{t=1}^T \Big(
  \describe{\E{\q(s_t|o_{\leq t},a_{<t})}{\ln\p(o_t|s_t, {\color{red}c})}}{reconstruction} 
  \quad-\describe{\Ebelow[\big]{\q(s_{t-1}|o_{\leq t-1},a_{<t-1}, {\color{red}c})}{\KL{\q(s_t|o_{\leq t},a_{<t}, {\color{red}c})}{\p(s_t|s_{t-1},a_{t-1}, {\color{red}c})}}}{complexity} \Big).
\end{aligned}
\end{multline}}

We mainly extend the steps in \citealt{hafner2019learning} for constructing the lower bound with context (derivation in \Cref{sec:derivs}). The expectations in this objective can be optimized with gradient ascent on samples drawn from the encoder using the reparameterization trick (\cite{kingma-iclr14a}).

\paragraph{Models} We follow the Dreamer~(v3) algorithm's (\cite{hafner-arxiv23a}) choice to split each latent state $s_t$ into a deterministic state $h_t$ and a stochastic state $z_t$.  This defines the cRSSM model (see \Cref{fig:cRssm}), which can be split into the following models:

\begin{gather}
\begin{aligned}
\makebox[12em][l]{Deterministic state model:} && h_t &=f_{\theta}(h_{t-1},z_{t-1},a_{t-1}, c) \\
\makebox[12em][l]{Stochastic state model:} && \hat{z}_t &\sim\pr[p_\theta](\hat{z}_t|h_t) \\
\makebox[12em][l]{Encoder} && z_t &\sim\pr[q_\theta](z_t|h_t, o_t) \\ 
\makebox[12em][l]{Observation model:} && \hat{o}_t &\sim\pr[p_\theta](\hat{o}_t|h_t,z_t, c) \\
\makebox[12em][l]{Reward model:} && \hat{r} &\sim\pr[p_\theta](\hat{r}_t|h_t,z_t,c) \\
\makebox[12em][l]{Continue model:} && \hat{n} &\sim\pr[p_\theta](\hat{n}_t|h_t,z_t,c).
\label{eq:c-rssm}
\end{aligned}
\end{gather}%
Refer to \Cref{sec:intutive_crssm} for an intutive explanation of how the RSSM and cRSSM work.

\paragraph{Parameterizing the Models} We do not change Dreamer's neural network architecture choices to parameterize these models. 
To train the objective, the Dreamer algorithm uses the past experiences of the agent (an actor-critic policy), which is trained concurrently with the cRSSM.

\subsubsection{Dreaming of Many Worlds}
Starting from a state $s_\tau$, inferred at some timestep $\tau$ from an observation sequence $o_{1:\tau}$ and actions $a_{1:\tau-1}$ and the true or factual context $c_{F}$ for that sequence, we can use the cRSSM to sample trajectories in the latent state space.

The cRSSM also allows for imagining trajectories for counterfactual contexts, or ``dreaming of many worlds''. We can do so by switching the context $c_{F}$, which governs the episode where the observations ($o_{1:\tau}$) used to infer the start state of the imagination $s_\tau$ were generated and dreaming further from that point in a different counterfactual context $c_{CF}$.

\subsubsection{Actor-Critic Policy}

We largely follow Dreamer(v3) regarding training an actor-critic policy on imagined trajectories. However, we introduce context into policy learning by conditioning the actor and critic networks with the context. The actor is optimized to maximize the expected return on the imagined trajectories.

\begin{equation}
\begin{aligned}
\makebox[12em][l]{Actor:} && &a_\tau \sim\ \pr[\pi_\phi](a_\tau|s_\tau, c) \\
\makebox[12em][l]{Critic:} && &v_\psi(s_t, c) \approx \E{\pi(\cdot|s_\tau, c)}{\textstyle\sum_{\tau=0}^{H-t}\gamma^{\tau}r_{t+\tau}}.
\label{eq:action_value_model}
\end{aligned}
\end{equation}%
 
\subsection{Naive Use of Context in Dreamer}
As discussed previously, Dreamers ZSG capabilities have not been explored in an observable context setting, nor, to the best of our knowledge, has this setting been explored in MBRL in general. Thus, we propose and discuss naive learning variants from and within the contextual setting and contrast them with our proposed cRSSM. The naive variants are then used as baselines in our experiments.
\subsubsection{Context as an Observation}
A commonly adopted approach to incorporate the inferred or true context into an algorithm is to concatenate it with the state or observation \citep{perez2020generalized,biedenkapp-prl22a,sodhani2022block,benjamins-tmlr23a}. 
We study applying this approach to vanilla Dreamer. Vanilla Dreamer optimizes all the objectives defined in \Cref{subsec:Dreamer}, but it does so without incorporating context in those objectives. To incorporate context naively, we provide it concatenated with the observation to the stochastic state encoder.  $\pr[q_\theta](z_t|h_t, x_t)$ where $x_t \dot = [ c_t, o_t ]$. Note that only the encoder gets the observation (or here the observation with context) as an input. The decoder then has to learn to reconstruct the context as it is part of the observation. The latent dynamics predictor used for imagination does not condition on observations. For a consistent imagination, the RSSM is burdened with retaining the context value (provided as an observation) which got encoded into the latent state, from which imagination begins. Since we only provide the latent state inferred by the encoder to the actor-critic model in this setting, if the context is not retained, then the actor-critic network will also not have access to the context information to learn accurate policies. This could make it hard to generalize OOD.
Still, the simplicity of directly using context as part of the observation is appealing, which has led to it being the predominant approach in model-free cRL.

\subsubsection{Hidden Context}

The cMDP is a sub-class of POMDP (\cite{kirk-jair23a}). As the vanilla Dreamer algorithm applies to POMDPs, we can use it without modification in the cRL setting without providing the context but training on episodes drawn from a training distribution over contexts. 

This is similar to the \emph{domain randomization} \citep{tobin2017domain, peng2018sim2real, andrychowicz2020learning} where the aim is to train the policy on a context distribution, usually inside the simulator, to aid generalization to some target context, usually on the real world. While applying domain randomization, most approaches aim to cover the target context distribution upon which they aim their policies to generalize. While studying ZSG, we also care about OOD contexts.


Unlike context-unaware model-free domain randomization approaches that learn representations purely from the reward signal, Dreamer's world modeling objective provides a useful inductive bias that could allow the model's observation encoder to learn how to infer the context implicitly more efficiently. 
With the clear disadvantage of not using context information when it is available, this approach might not be able to learn to distinguish which exact context setting it properly is dealing with. Consequently, the resulting policies might act for a spurious context and thus behave sub-optimally or even fail catastrophically.
However, providing a training distribution of contexts might already be enough for the world model to infer the context, especially if the context is implicitly encoded in the observations (e.g., the pole length in CartPole with pixel observations). Thus, this style of context handling can be viewed as a simple context-inference approach.
In this setting, similar to treating context as an observation, we only use the latent state inferred by the encoder as the input to the policy network. 

\section{Experiments and Discussion}

In this section, we assess the performance of Dreamer in achieving generalization under observed contexts. We compare various context conditioning approaches, following the evaluation protocols outlined in \citep{kirk-jair23a}. In particular, our findings highlight the effectiveness of our cRSSM method, showcasing quantitative and qualitative results in terms of zero-shot generalization (ZSG), particularly in scenarios involving changes to context parameters affecting the dynamics.

\paragraph{Environments:}

Our experiments leverage CARL, tailored for our investigation into ZSG. In CARL we pick the following environments and contexts.
\begin{itemize}\itemsep0em
    \item \textbf{CartPole} \citep{barto1983neuronlike}: \texttt{pole length} and \texttt{gravity}.
    \item \textbf{DMC Walker Walk} \citep{tassa2018deepmind}: \texttt{actuator strength} and \texttt{gravity}.
\end{itemize}

As these environments are good examples of the desired continuous contextual settings, they provide a suitable benchmark for our study. CartPole serves as a simple problem, while DMC Walker presents a more complex challenge. This approach aligns with prior works, such as \cite{sodhani2022block} and \cite{zhang-iclr21a}, which evaluate generalization on DMControl tasks to changes in context.

For each environment, we use two modalities of observation, namely
\begin{inparaenum}[(1)]
    \item \textbf{Featurized:} This uses featurized observations, which are generally easier to learn policies as they exhibit the least or no partial observability depending on the environment;
    \item \textbf{Pixel:} Image observations which are more difficult as the model has to infer the latent states from it.
\end{inparaenum}

\paragraph{Training Pipeline}

We use Dreamer~(v3) default hyperparameters for all experiments, with $50$k steps for CartPole and $500$k for DMC Walker (10 seeds). We also show DMC Walker results with 10 seeds and $100$k steps in \Cref{sec:walker_less_samples} to analyze performance with fewer samples. Refer \Cref{app:hparam} for exact hyperparameter values.

The CARL benchmark provides default context values (i.e., those commonly used in the literature for single-environment training) for the two context dimensions we consider for each environment. For each context type, we define \textit{training} and \textit{evaluation} ranges. The default value, training ranges, and evaluation context values are provided in \Cref{sec:context_range}.
We train our three methods, namely the \emph{\ccrssm}, \emph{\cconcatagent} and \emph{\chiddenagent}, in the following  \textit{training settings}:
\begin{enumerate}\itemsep0pt
    \item \textbf{Single context variation}: We sample $100$ context values uniformly for one context dimension in its training range, keeping the other fixed to its default value and vice versa.
    \item \textbf{Dual context variation}: We sample $100$ context values uniformly in the training range of both context dimensions.
\end{enumerate}

\begin{figure}\vskip-.2in
    \centering
    \includegraphics{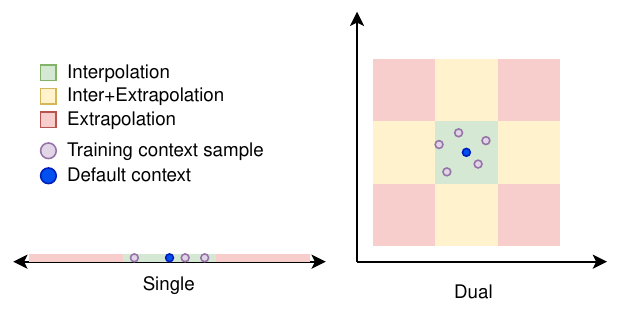}\vskip-.15in
    \caption{Training contexts and evaluation regions for single and dual context variation.}
    \label{fig:context-protocol}
\end{figure}

\paragraph{Evaluation Protocol} Following \citet{kirk-jair23a} we evaluate our agents on the following \textit{evaluation settings} (visualized in \cref{fig:context-protocol}):
\begin{enumerate}\itemsep0pt
    \item \textbf{Interpolation (I)}: Evaluation contexts are selected fully within the training range.
    \item \textbf{Inter+Extrapolation (I+E)}: Evaluation contexts are selected to be within the training distribution for one context dimension and out-of-distribution (OOD) for another. This evaluation setting only applies to agents trained in the dual context variation setting. 
    \item \textbf{Extrapolation (E)}: Evaluation contexts are fully OOD, as they are selected outside the training context set limits.
\end{enumerate}

To gain insights into Dreamers' basic generalization capabilities, we also train context-unaware agents on the single default context (\emph{\cdefaultagent}) per environment. We evaluate these hidden context default agents in the same context values used for the three evaluation settings to compare each of our methods trained in the two training settings. 

In the evaluation protocol, each context could constitute a task of different difficulty. For example, learning to control an agent in a context where it carries lighter loads might be easier than the one with heavier loads. Following \cite{benjamins-tmlr23a}, to obtain an upper bound of the policy returns, we train \emph{\expertpol agents} for selected contexts that broadly cover our ranges of training and evaluation contexts. Expert agent performances are the best mean return over $50$ episodes among five seeds. This gives an approximate upper bound on the returns achievable if Dreamer is trained in a particular context. We used the best \emph{\randompol policy} mean return on $50$ episodes over five seeds to define a lower bound. See \Cref{subsec:expert_perf} for the detailed expert and random agents performance.

\paragraph{Evaluation Metric}
We use the performance of the \expertpol and \randompol policies to normalize our evaluation performance. A normalized score of $1.0$ would indicate the expert performance of an agent in that setting, and $0.0$ performance equal to a random policy. Since we evaluate our approaches on more contexts than the number of experts, we pick the nearest context (normalized to account for different scales of contexts) for which an expert is available and use it as reference.

Following the recommendation of \citet{agarwal-neurips21a}, we report the interquartile mean (IQM) of the normalized aggregated scores across contexts in different regions for each experiment.

\begin{figure}[!htb]
    \centering
    \includegraphics[width=\textwidth]{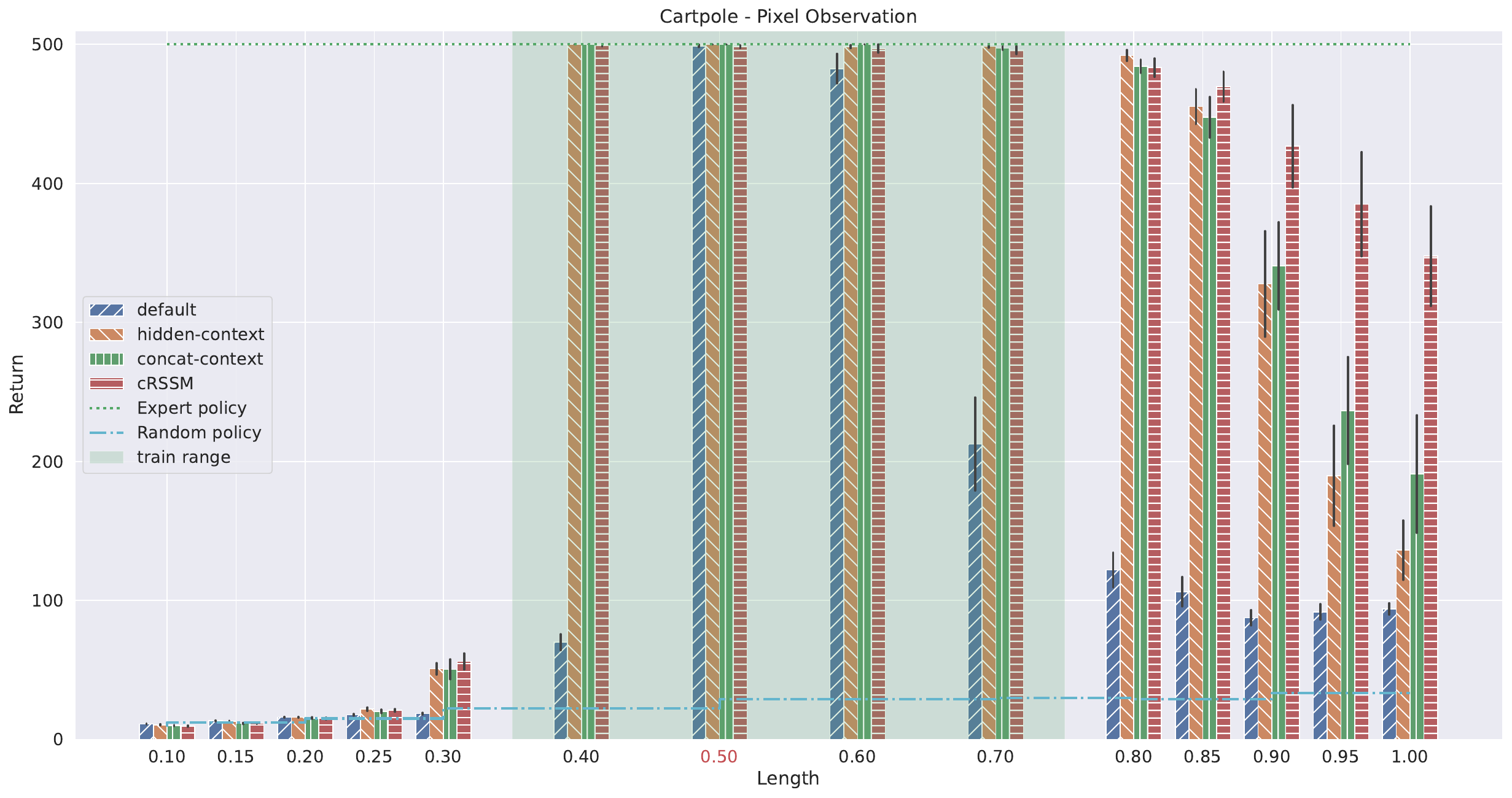}
    
    \caption{Generalization capabilities of Dreamer with pixel-observations when varying the pole length in CartPole.
    The y-axis indicates the gained reward, and the x-axis the pole length.
    The {\color{default}blue} bars shows vanilla Dreamers performance when only training on the default length ($0.50$) and extrapolating to other settings. The shaded are gives the training range for the methods using context. {\color{ep}Expert} and {\color{rp}random} policies give upper and lower bound for performances in each context.
    }
    \label{fig:exp-dreamer-CartPole-length}
\end{figure}

\begin{figure}[tbp]
    \centering
    \begin{subfigure}[b]{\textwidth} 
        \includegraphics[width=\textwidth]{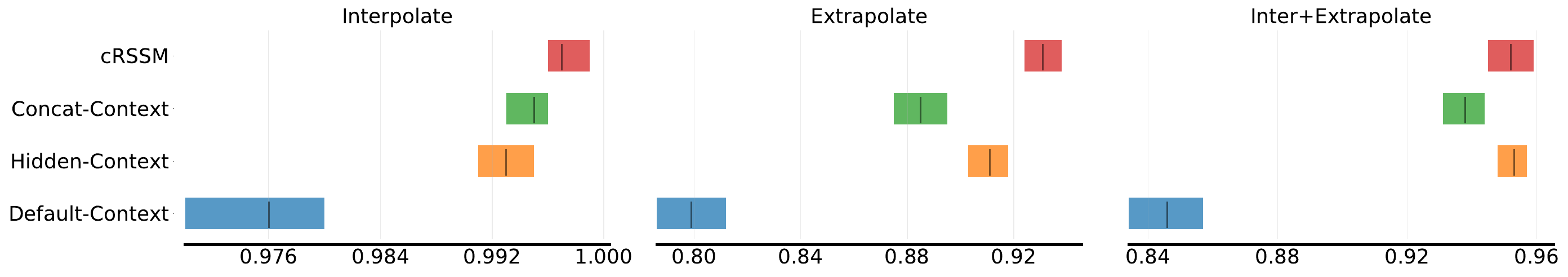}%
        \caption{Feature based IQM}%
        \label{fig:feature_agg}
    \end{subfigure}
    \begin{subfigure}[b]{\textwidth} 
        \includegraphics[width=\textwidth]{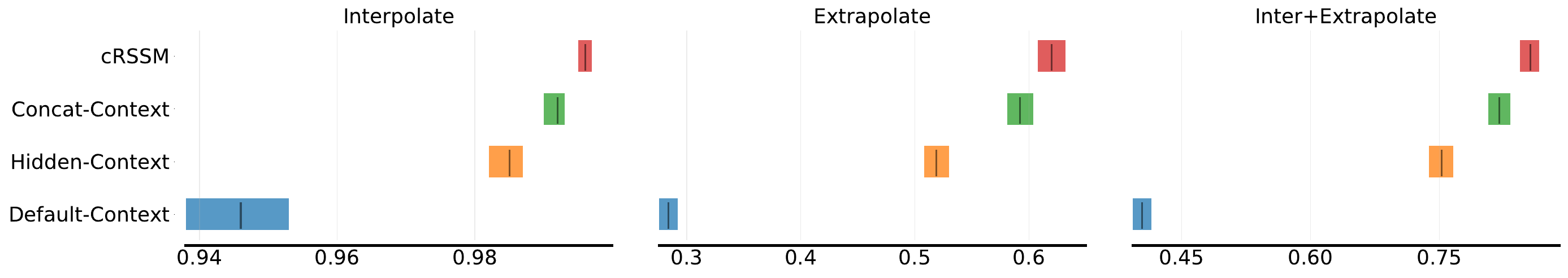}%
        \caption{Pixel based IQM}%
        \label{fig:pixel_agg}
    \end{subfigure}
   
    \caption{Aggregated comparison (across contexts \& tasks) across \emph{\ccrssm}, \emph{\cconcatagent}, \emph{\chiddenagent}, and \emph{\cdefaultagent} for the evaluation settings: Interpolation, Extrapolation, and Inter+Extrapolation using IQM over expert normalized scores for both input modalities. Intervals shown are stratified bootstrap 95\% confidence intervals over seeds \& aggregated contexts.}
    \label{fig:aggregate_performance_iqm}
\end{figure}

\begin{table}[t]
\caption{Results for different evaluation settings, in featurized and pixel modality. Each described by three variables: context, method, and mode. Context takes values from $\{c_1, c_2\}$ where $c_1$ is gravity for both CartPole and Walker environments, $c_2$ is pole length for CartPole and Actuator Strength for Walker. $c_{1+2}$ indicating multiple contexts; and method from $\{ {\color{default}d: \text{{\defaultagent}}}, {\color{hidden}h: \text{{\hiddenagent}}}, {\color{concat}c: \text{{\concatagent}}}, {\color{crssm}cR: \text{{\crssm}}} \}$}%
    \label{tab:quantitative_results}%
    \setlength\tabcolsep{3pt}
    \centering
    \begin{tabular}{lccc|ccc|ccc|ccc}
        \toprule
        \multicolumn{1}{c}{Setting} & I & E & I+E  & I & E & I+E & I & E & I+E & I & E & I+E  \\ \midrule
        \multicolumn{1}{c}{} & \multicolumn{6}{c}{CartPole} & \multicolumn{6}{c}{Walker}  \\ 
        \cmidrule(lr){2-7}  \cmidrule(lr){8-13} 
                            &    \multicolumn{3}{c}{Featurized} &  \multicolumn{3}{c}{Pixel}  \\ 
        \cmidrule(lr){2-7} \cmidrule(lr){8-13} 
        ($c_1$ {\color{default}d}) & 1.000 & 1.000 & - & 1.000 & 0.938 & - & 0.903 & 0.561 & - & 0.940 & 0.546 & - \\
        ($c_1$ {\color{hidden}h}) & 1.000 & 1.000 & - & 1.000 &  0.995 & - & 0.967 & 0.764 & - & 0.945 & 0.708 & - \\
        ($c_1$ {\color{concat}c}) & 1.000 & 1.000 & - & 1.000 &  0.997 & - & 0.966 & 0.769 & - & \textbf{0.966} & \textbf{0.733} & - \\
        ($c_1$ {\color{crssm}cR}) & 1.000 & 1.000 & - & 1.000& \textbf{1.000} & - & \textbf{0.985} & \textbf{0.806} & - & 0.959 & 0.710 & - \\
         \midrule
        ($c_2$ {\color{default}d}) & 1.000 & 0.995 & - & 0.677 & 0.059 & - & 0.885 & 0.479 & - & 0.959 & 0.461 & - \\
        ($c_2$ {\color{hidden}h}) & 1.000 & 0.996 & - & \textbf{1.000} & 0.169 & - & 0.959 & 0.571 & - & 0.947 & 0.571 & - \\
        ($c_2$ {\color{concat}c}) & 1.000 & 0.987 & - & \textbf{1.000} & 0.210 & - & 0.926 & 0.597 &- & 0.983 & \textbf{0.635} & - \\
        ($c_2$ {\color{crssm}cR}) & 1.000 & \textbf{1.000 }& - & \textbf{1.000} & \textbf{0.374} & - & \textbf{0.998} & \textbf{0.674} & - & \textbf{0.994} & 0.623 & - \\
        \midrule
        ($c_{1+2}$ {\color{default}d}) & 1.000 & 0.945 & 0.998 & 0.901 & 0.038 & 0.210 & 0.842 & 0.520 & 0.595 & 0.915 & 0.503 & 0.570 \\
        ($c_{1+2}$ {\color{hidden}h}) & 1.000 & 0.989 & \textbf{1.000} & \textbf{1.000} & 0.149 & 0.701 & 0.966 & \textbf{0.764} & \textbf{0.843} & 0.952 & 0.666 & 0.772 \\
        ($c_{1+2}$ {\color{concat}c}) & 1.000 & 0.970 & \textbf{1.000} & \textbf{1.000} & 0.257 & 0.779 & 0.972 & 0.727 & 0.830 & 0.965 & \textbf{0.724} & 0.823 \\
        ($c_{1+2}$ {\color{crssm}cR}) & 1.000 & \textbf{0.997} & \textbf{1.000} & \textbf{1.000} & \textbf{0.334} & \textbf{0.826} & \textbf{0.982} & 0.677 & 0.820 & \textbf{0.988} & 0.691 & \textbf{0.843}  \\
        \bottomrule
    \end{tabular}%

\end{table}%

\subsection{Results}

In this section, we analyze our results to answer three key research questions that motivate our study of ZSG and our method of context conditioning. To help answer the first two questions, we first provide the results on the representative Cartpole with pixel observations setting, comparing the mean evaluation returns across our methods in \Cref{fig:exp-dreamer-CartPole-length}. To compare the overall performance of our different modalities for our four methods, in \Cref{fig:aggregate_performance_iqm}, we provide the aggregated IQM over normalized return along with stratified bootstrap 95\% confidence intervals \cite{agarwal-neurips21a} across different contexts, single/dual-variation training paradigms, and environments. We also present the aggregated probability of improvement for \crssm compared to other methods in \Cref{subsec:POI}. Individual results comparing the raw returns of all agents in different contexts, modalities, and context variation settings are available in \Cref{app:perfs}.
We also report the aggregated IQM scores for different context regions for each individual setting in  \Cref{tab:quantitative_results}. Refer \Cref{sec:individual_icm} plots for these individual IQMs and 95\% confidence intervals.

\subsubsection{How Effective is Domain Randomization for Dreamer's ZSG?}

To answer this, we compare the two approaches \emph{\cdefaultagent}, with the agent trained on the default context $c_d$, and the \emph{\chiddenagent}, which involves training with domain randomization of contexts.

As a motivating example to compare these methods, we first present a representative result in \Cref{fig:exp-dreamer-CartPole-length} for the different methods trained on the Cartpole environment with pixel observations and varying the pole length. We observe that the \emph{\hiddenagent} agent significantly outperforms the \textit{\defaultagent} agent, especially in the extrapolation setting. The performance of \textit{\defaultagent} agent drops noticeably when it moves away from its familiar default context. The aggregated results across the interpolation and extrapolation regime for this setting are available in the pixel column under the rows (l {\color{default}d}/{\color{hidden}h}/{\color{concat}c}/{\color{crssm}cR}) for the Cartpole group in \Cref{tab:quantitative_results}.

The aggregated metrics in \Cref{fig:aggregate_performance_iqm} show that \emph{\hiddenagent} outperforms \emph{\defaultagent} in all settings. The improvement is more pronounced in the pixel-based modality (\Cref{fig:pixel_agg}). This highlights the impact of domain randomization for generalization to unseen contexts in the more complex pixel modality, as this exhibits increased partial observability.
 
In summary, domain randomization benefits the ZSG of the Dreamer algorithm, and the improvement is striking for the pronounced pixel modality on the evaluated tasks.

\subsubsection{Does Explicit Context Conditioning Aid ZSG?} 

Having established the benefits of domain randomization through \emph{\chiddenagent} for Dreamer's ZSG on the given tasks, our focus shifts to evaluating the impact of explicit context conditioning methods, namely \emph{\ccrssm}, our principled way to incorporate context into Dreamer's world model; and \emph{\cconcatagent} where we augment the observations with the context.

In the Cartpole environment, for the pixel modality observations (\Cref{fig:exp-dreamer-CartPole-length}), both explicit conditioning methods, \emph{\crssm} and \emph{\concatagent}, demonstrate superior performance over \emph{\hiddenagent}, particularly in scenarios with longer pole lengths. Here, \emph{\crssm} emerges as the frontrunner.

To extend this analysis to all of our settings, we again turn to the aggregated IQM scores. For the featurized modality (\Cref{fig:feature_agg}), the \emph{\crssm} significantly outperforms both \emph{\hiddenagent} and \emph{\concatagent}. In contrast, \emph{\concatagent} trails behind \emph{\hiddenagent} in \emph{inter+extrapolation} and extrapolation settings. In the more challenging pixel modality (refer to \Cref{fig:pixel_agg}), explicit context conditioning techniques demonstrate significant improvements over the \emph{\hiddenagent} across all evaluation scenarios, highlighting the importance of context conditioning for generalization.

Between the explicit context conditioning methods, \crssm performs best in all evaluation regions on aggregate. The improvements are particularly pronounced in the more challenging extrapolation and \emph{inter+extrapolation} scenarios. Following \citet{agarwal-neurips21a}, we provide the probability of improvement of cRSSM over other methods in \Cref{subsec:POI}, solidifying our claims.

For a detailed breakdown of each task, context variation, and evaluation protocol, we consult \Cref{tab:quantitative_results}. In the Cartpole environment, in the featurized case, all methods perform similarly in all settings and evaluation regions. In dual context variation, \defaultagent lags behind other approaches which shows benefit of varying context during training even in this simple setting. In the pixel modality the differences among methods are most discernible, context conditioning methods outperform domain the \emph{\hiddenagent}. And among context-conditioning the \emph{\crssm} outperforms \emph{\concatagent} context, particularly excelling in variations of pole length and combinations of length and gravity.

In the DMC Walker environment, context conditioning methods perform better than \emph{\hiddenagent}.

Within the featurized category, \emph{\crssm} takes the top spot. However, in the pixel modality, \emph{\concatagent} leads, with \emph{\crssm} slightly behind, except for the inter+extrapolation setting where \emph{\crssm} demonstrates a better understanding of the meaning of each context separately.

In summary, explicit context conditioning aids ZSG. \emph{\crssm} showcases improved generalization across all modalities in the Cartpole environment and delivers substantial generalization improvements in the featurized modality of DMC Walker, albeit lagging slightly behind in the pixel modality.

\subsubsection{Beyond Task Performance, Does Context-Conditioning Impact the Latent States?}

We qualitatively assess the ability of different methods to understand context and its impact on latent representations to help explain differences in performance and shortcomings of different approaches. 

\begin{figure}[!htb]
    \centering
    \begin{minipage}[b]{0.49\textwidth}
        \centering
        \begin{minipage}[b]{0.49\textwidth}\centering
            \begin{minipage}[b]{0.3\textwidth}
                \centering
                \includegraphics[width=0.99\textwidth]{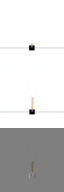}
                \subcaption*{Concat}
            \end{minipage}
            \begin{minipage}[b]{0.3\textwidth}
                \centering
                \includegraphics[width=0.99\textwidth]{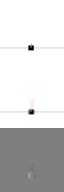}
                \subcaption*{cRSSM}
            \end{minipage}
            \subcaption*{Obs: short \\ Context: short}
        \end{minipage}
        \begin{minipage}[b]{0.49\textwidth}\centering
            \begin{minipage}[b]{0.3\textwidth}
                \centering
                \includegraphics[width=0.99\textwidth]{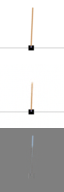}
                \subcaption*{Concat}
            \end{minipage}
            \begin{minipage}[b]{0.3\textwidth}
                \centering
                \includegraphics[width=0.99\textwidth]{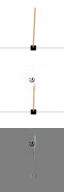}
                \subcaption*{cRSSM}
            \end{minipage}
            \subcaption*{Obs: long \\ Context: long}
        \end{minipage}
        \subcaption{Extrapolated Contexts}
        \label{extrapolate_context}
    \end{minipage} \vline
    \begin{minipage}[b]{0.49\textwidth}
        \centering
        \begin{minipage}[b]{0.49\textwidth}\centering
            \begin{minipage}[b]{0.3\textwidth}
                \centering
                \includegraphics[width=0.99\textwidth]{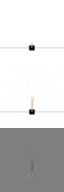}
                \subcaption*{Concat}
            \end{minipage}
            \begin{minipage}[b]{0.3\textwidth}
                \centering
                \includegraphics[width=0.99\textwidth]{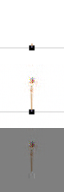}
                \subcaption*{cRSSM}
            \end{minipage} 
            \subcaption*{Obs: short \\ Context: long}
        \end{minipage}
        \begin{minipage}[b]{0.49\textwidth}\centering
            \begin{minipage}[b]{0.3\textwidth}
                \centering
                \includegraphics[width=0.99\textwidth]{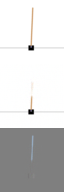}
                \subcaption*{Concat}
            \end{minipage}
            \begin{minipage}[b]{0.3\textwidth}
                \centering
                \includegraphics[width=0.99\textwidth]{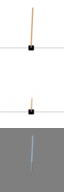}
                \subcaption*{cRSSM}
            \end{minipage}
            \subcaption*{Obs: long \\ Context: short}
        \end{minipage}
        \subcaption{Counterfactual Contexts}
        \label{counterfactual_context}
    \end{minipage} 
    \caption{Qualitative results for the model generative ability of novel context. In each image, we have the true observation, followed by the one reconstructed by the decoder with context conditioning from the latent encoded from the true image, and lastly the difference between the two images. The \textit{short} refers to a length of $0.1$ units, and \textit{long} is the OOD length of $1.0$ units. In the extrapolation case, ideally, the difference should be minimal and, in the counterfactual case maximal.}
    \label{fig:fourgroups}
\end{figure}

To evaluate how the methods use context, we visually investigate Dreamer's reconstruction of visually observable OOD contexts, which are either encoded into the latent state in \cconcatagent~or conditioned separately \ccrssm. We choose the visually observable context to be the pole length in Cartpole, evaluating lengths shorter and longer than what it has seen in the training distribution (labeled short and long in Figure~\ref{fig:fourgroups}). A model capable of generating images conditioned on novel context demonstrates a semantic understanding of the context that is well grounded in the image space. In cases involving context-conditioned models, we also provide counterfactual visual explanations, exploring how the reconstructed pixel observation $\hat{o}_t$ inferred from the original observation that encodes the factual context ($c_F$) differs if conditioned on a counterfactual context ($c_{CF}$).

\paragraph{Extrapolation}
In the extrapolation case depicted in \Cref{extrapolate_context}, we encode observations from the OOD pole length context (short: 0.1 or long: 1.0) and also condition on the true OOD context value. It can be seen that the \crssm~predicted observation is more faithful to the OOD contexts compared to other methods.  In the case of short length, \crssm~generates a slightly blurred and shorter pole, while \concatagent~is confined to the shortest in-distribution pole. 
For the longer pole, \crssm exhibits more realistic behavior by attempting to add additional pixels on top of the longest pole it has seen, demonstrating a better semantic understanding of the context by the world model. In contrast, \concatagent~ falls short and instead decodes the longest in-distribution pole length. We attribute the bottleneck that confines \concatagent~to the bounds of the context seen during training to the discretization of the latent states (containing the context).  Notably, the decoder's ability to decode shorter-length poles for shorter inputs and vice versa suggests that the encoder of both methods has learned the scale of pole lengths meaningfully.

\paragraph{Why Can Assessing Disentanglement Help?} Although we see a meaningful extrapolation, the length and pole position could also have been encoded into the latent state. This would defeat the purpose of conditioning on the context. Ideally, we want the system to interpret our context as the source of truth and not redundantly encode it from observations. The ``sparse mechanism shift'' hypothesis \citep{schölkopf2021towards} states that such disentanglement of causal mechanisms in representations enables OOD generalization.

\paragraph{Counterfactual Assessment of Disentanglement} To test for disentanglement and faithfulness to the conditioning context, we use the ability of our \crssm world model to dream of many worlds by taking observations generated from the factual context $c_F$ and encoding them to the latent state while conditioning the model in the counterfactual context $c_{CF}$. Then, we decode the image to see how counterfactual conditioning influences image generation. From \Cref{counterfactual_context}, we can see that \crssm uses the context value more faithfully than ~\concatagent. This demonstrates the capability of cRSSM to extrapolate and combine the conditioning context with the latent state to generate semantically meaningful counterfactual images. In contrast to the context-disentangled latent space of cRSSM, the \concatagent approach encodes both context information and observations jointly into the latent state, hindering its ability to generalize effectively.

Our investigation reveals clear evidence of extrapolation capabilities in our proposed principled \crssm~approach compared to the vanilla \concatagent~strategy. Furthermore, through our visual counterfactual explanations, we observe indications that the latent state in \crssm~appears disentangled from the context, which explains the observed gains in generalization using this approach.

\section{Conclusion and Future Work}
We studied zero-shot generalization in Dreamer-style model-based reinforcement learning through the lens of contextual reinforcement learning. We discussed naive ways to incorporate contextual information into the MBRL learning pipeline and formulated the novel \crssm for Dreamer.
Our \crssm provides a systematic approach to using context in the world modelling objectives under partial observability. 
Our experiments, using a rigorous evaluation protocol for zero-shot generalization, showed that naive approaches, such as domain randomization improve generalization performance. However, more principled methods such as our \crssm are required to perform significantly better in-distribution and out-of-distribution. Our study opens the door to future work on zero-shot generalization for MBRL approaches such as Dreamer. Creating contextual benchmarks for environments such as Atari \citep{bellemare2013arcade}, DMLab \citep{Beattie2016DeepMindL}, ProcGen \citep{anand2021procedural}, and Minecraft \citep{guss2019minerl} would be an interesting avenue for future research into ZSG, as contextual changes in these benchmarks would necessitate more pronounced changes in policies. We discuss this in more detail in \Cref{app:benchmarks}.

Our current \crssm formulation assumes that context is observable, meaning it is directly available as input. We plan to extend the \crssm formulation to cases where context is not directly observable and must be inferred along with the latent states. While we show qualitative results for counterfactual dreams, the next step would be to use this to generate dreams for counterfactual contexts during training and study the effect on ZSG and sample efficiency.


\subsubsection*{Acknowledgments}
\label{sec:ack}
S.P. and K.F. acknowledge funding by the Konrad Zuse School of Excellence in Learning and Intelligent Systems (ELIZA) grant. R.R. and A.B. acknowledge funding from the Carl Zeiss Foundation through the research network ``Responsive and Scalable Learning for Robots Assisting Humans'' (ReScaLe) of the University of Freiburg.

\bibliography{bib/strings,bib/local,bib/lib,bib/proc}
\bibliographystyle{rlc}

\clearpage
\appendix




\section{cRSSM Bound Derivation}
\label{sec:derivs}
The variational bound for contextual latent dynamics models $\p(o_{1:H},s_{1:H}|a_{1:H}, c)=\prod_t\p(s_t|s_{t-1},a_{t-1}, c)\p(o_t|s_t, c)$ and a variational posterior $\q(s_{1:H}|o_{1:H},a_{1:H}, c)=\prod_t\q(s_t|o_{\leq t},a_{<t}, c)$ follows from importance weighting and Jensen's inequality as shown,
\begin{equation}
\begin{aligned}
\ln\p(o_{1:H}|a_{1:H}, c)
&\triangleq\ln\E[\bigg]{\p(s_{1:H}|a_{1:H}, c)}{\prod_{t=1}^H \p(o_t|s_t, c)} \\
&=\ln\E[\bigg]{\q(s_{1:H}|o_{1:H},a_{1:H}, c)}{\prod_{t=1}^H \p(o_t|s_t, c)\p(s_t|s_{t-1},a_{t-1}, c)/\q(s_t|o_{\leq t},a_{<t}, c)} \\
&\geq\E[\bigg]{\q(s_{1:H}|o_{1:H},a_{1:H}, c)}{\sum_{t=1}^H \ln\p(o_t|s_t, c)+\ln\p(s_t|s_{t-1},a_{t-1}, c)-\ln\q(s_t|o_{\leq t},a_{<t}, c)} \\
&=\sum_{t=1}^H \Big(
  \describe{\Ebelow{\q(s_t|o_{\leq t},a_{<t}, c)}{\ln\p(o_t|s_t, c)}}{reconstruction}
  -\describe{\Ebelow[\big]{\q(s_{t-1|o_{\leq t-1},a_{<t-1}, c})}{\KL{\q(s_t|o_{\leq t},a_{<t}, c)}{\p(s_t|s_{t-1},a_{t-1}, c)}}}{complexity} \Big).
\end{aligned}
\label{eq:elbo_deriv}
\end{equation}%

\section{Train and Evaluation Context Ranges}
\label{sec:context_range}

\begin{table}[!htb]
\centering
\begin{tabularx}{\textwidth}{XXX}
\toprule
Context & Gravity & Length \\ \midrule
Default & 9.8 &  .5 \\
Training Range & [4.9, 14.7] & [.35, .75] \\
Single Evaluation Values &  .98, 17.15, 2.45, 3.92, 4.9, 7.35, 9.8, 12.25, 14.7, 15.68, 16.66, 17.64, 18.62, 19.6 & .1, .15, .2, .25, .3, .4, .5, .6, .7, .8, .85, .9, .95, 1.0 \\
Dual Evaluation Values & .98, 2.45, 3.92, 15.68, 17.64, 19.6 & .1,  .2, .3, .5, .7, .8, .9, 1.0 \\ \bottomrule
\end{tabularx}
\caption{CartPole Context Values}
\end{table}
\begin{table}[!htb]
\centering
\begin{tabularx}{\textwidth}{XXX}
\toprule
Context & Gravity & Actuator Strength \\ \midrule
Default & 9.8 &  .5 \\
Training Range & [4.9, 14.7] & [.5, 1.5] \\
Single Evaluation Values &  .98, 17.15, 2.45, 3.92, 4.9, 7.35, 9.8, 12.25, 14.7, 15.68, 16.66, 17.64, 18.62, 19.6 & .1, .2, .3, .4, .5, .75, 1.0, 1.25, 1.5, 1.6, 1.7, 1.8, 1.9, 2.0 \\
Dual Evaluation Values & .98, 2.45, 3.92, 15.68, 17.64, 19.6 & .1, .3, .5, 1.0, 1.5, 1.6, 1.8, 2.0 \\ \bottomrule
\end{tabularx}
\caption{DMC Walker Context Values}
\end{table}
\clearpage

\section{Agent Performances}\label{app:perfs}

\subsection{Expert and Random agent performance}
\label{subsec:expert_perf}
\begin{figure}[htb!]
    \centering
    \begin{subfigure}{0.45\textwidth}
        \includegraphics[width=\linewidth]{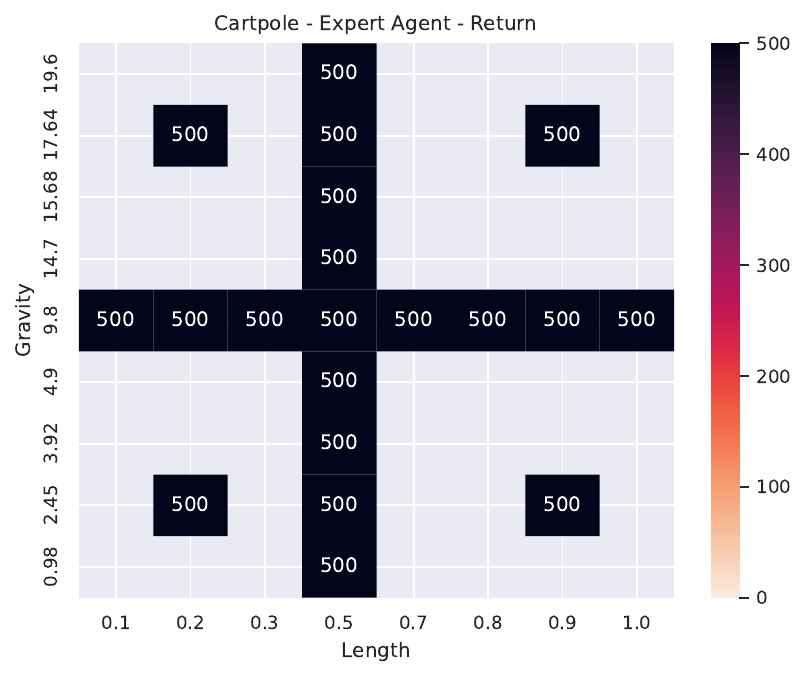}
    \end{subfigure}
    \begin{subfigure}{0.45\textwidth}
        \includegraphics[width=\linewidth]{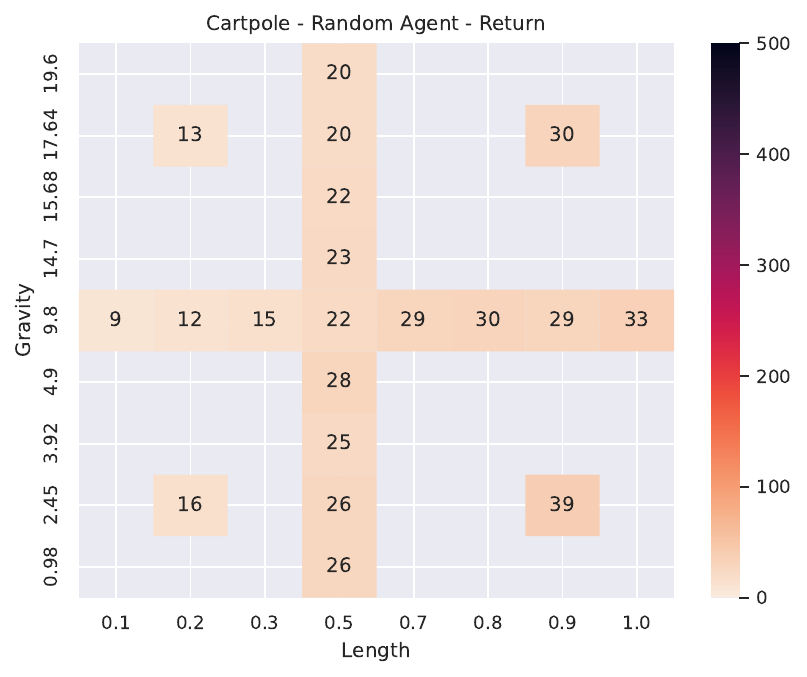}
    \end{subfigure}

    \medskip
    
    \begin{subfigure}{0.45\textwidth}
        \includegraphics[width=\linewidth]{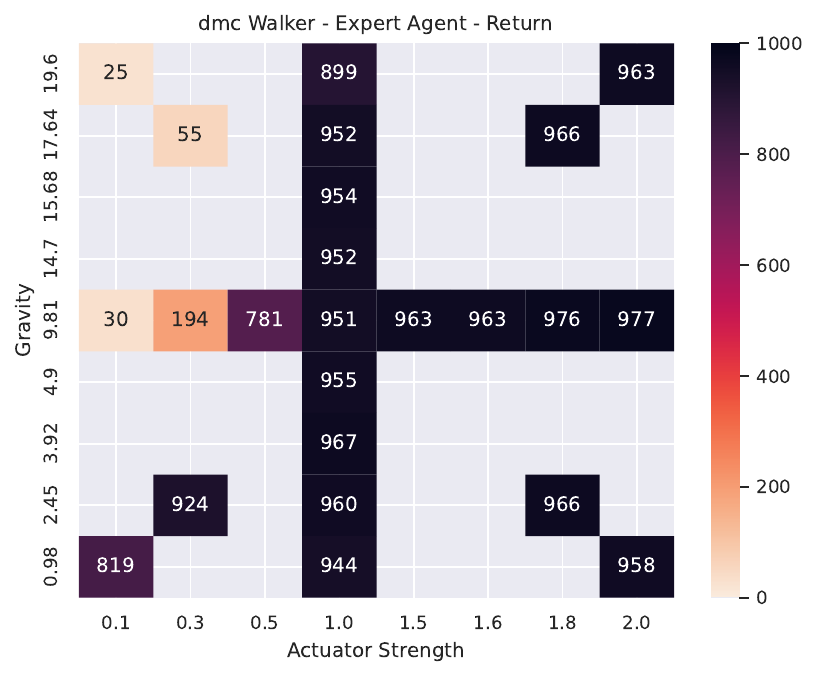}
    \end{subfigure}
    \begin{subfigure}{0.45\textwidth}
        \includegraphics[width=\linewidth]{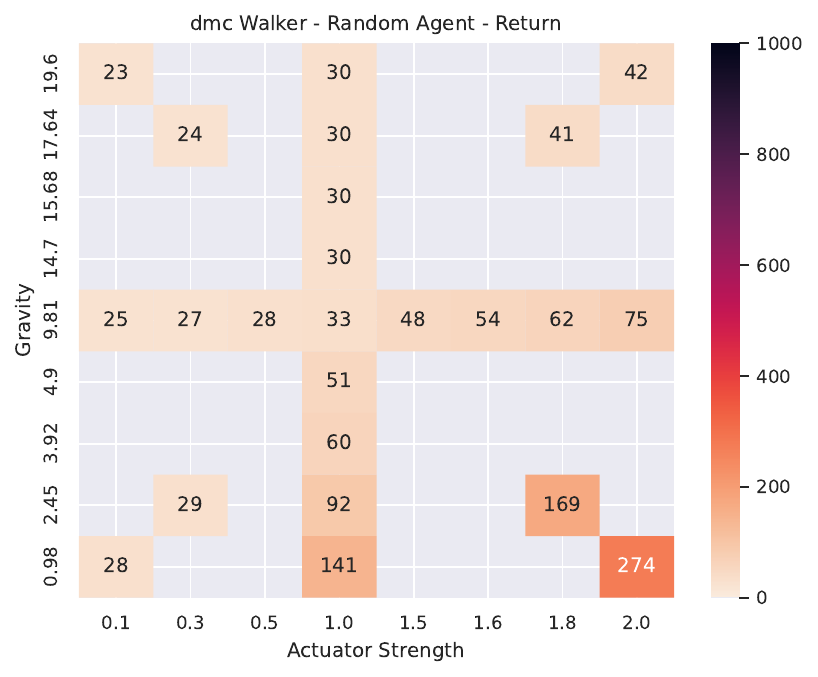}
    \end{subfigure}
    \caption{The best performing random policy and expert trained on each context over 5 seeds. We use featurized modality with less partial observability compared to pixels, to get an optimistic upper bound of expert returns.}
\end{figure}
\clearpage
\subsection{Varying single context}

\begin{figure}[htb!]
    \centering
    \begin{subfigure}{\textwidth}
        \includegraphics[width=\linewidth]{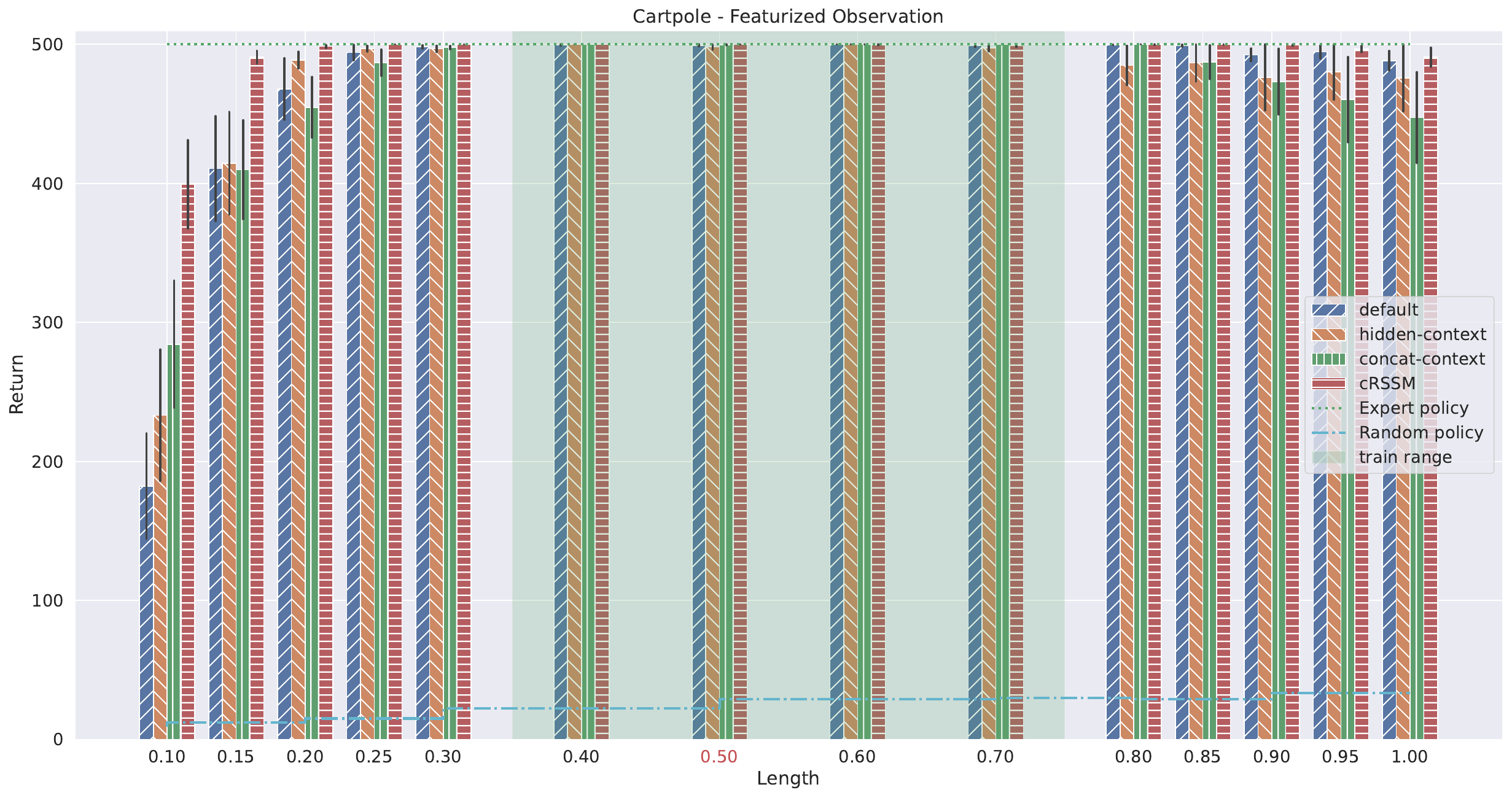}
    \end{subfigure}
    \medskip

    \begin{subfigure}{\textwidth}
        \includegraphics[width=\linewidth]{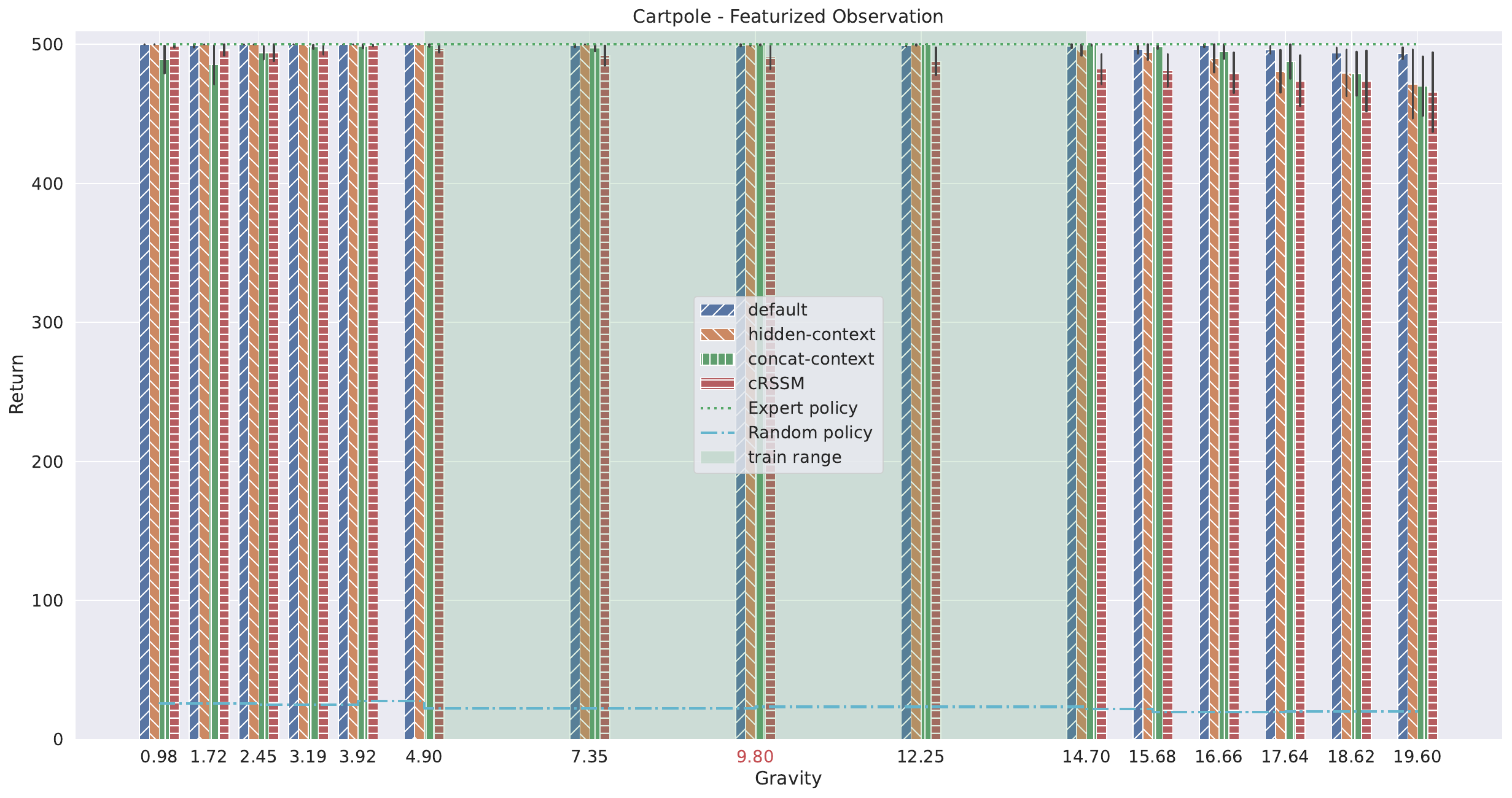}
    \end{subfigure}

    \medskip
    \caption{CartPole - Featurized Observations - The mean and standard error of the average evaluation returns are computed across 10 seeds, for 50 evaluation episodes each}\label{fig:cartpole-feat-appendix}
\end{figure}

\begin{figure}[htb!]
    \centering
    
    \begin{subfigure}{\textwidth}
        \includegraphics[width=\linewidth]{figures/return_plots/cartpole_length_pixel_observation_return.pdf}
    \end{subfigure}
    \begin{subfigure}{\textwidth}
        \includegraphics[width=\linewidth]{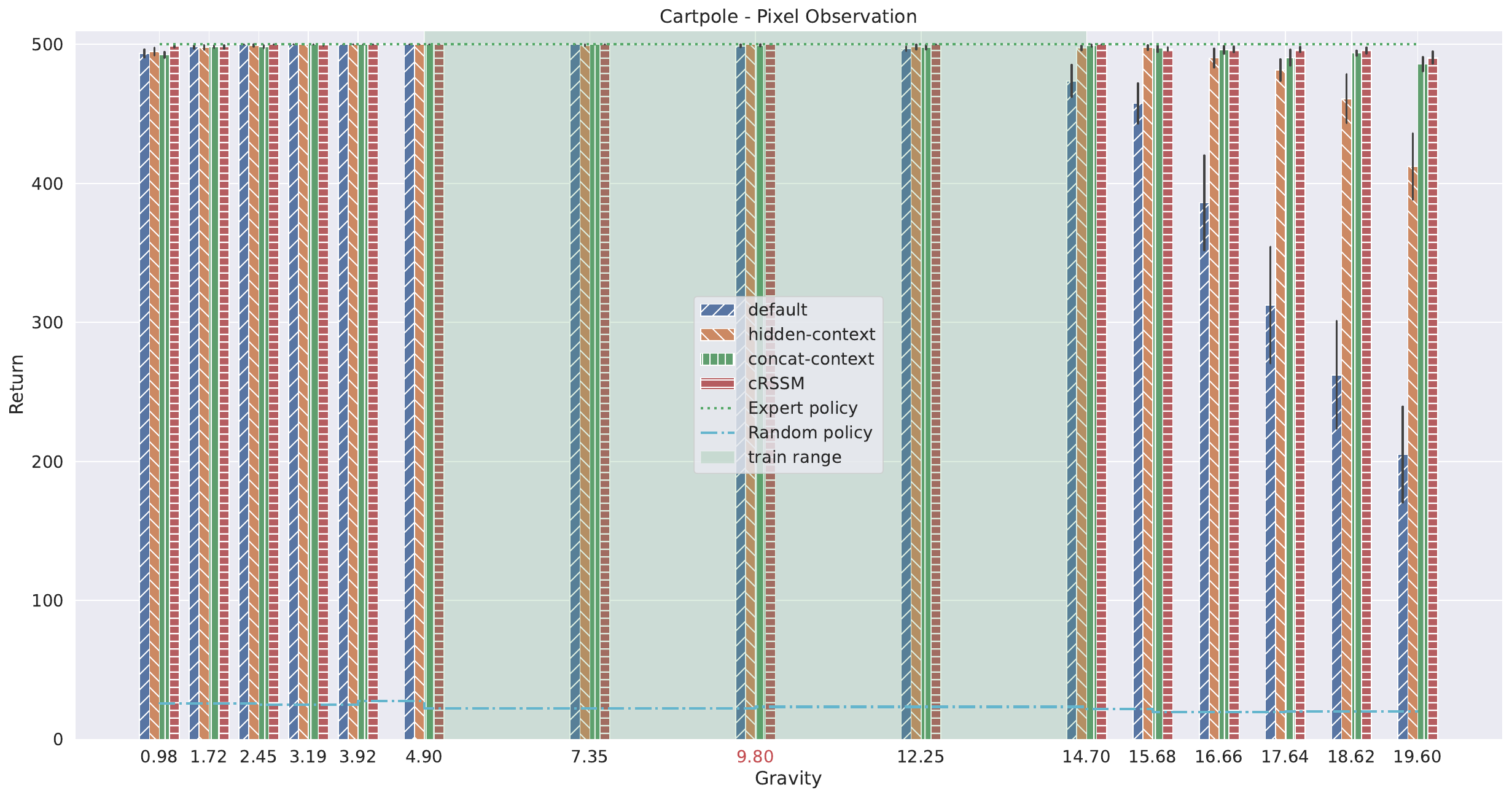}
    \end{subfigure}

    \medskip
    \caption{CartPole Pixel Observations - The mean and standard error of the average evaluation returns are computed across 10 seeds, for 50 evaluation episodes each}
\end{figure}

\begin{figure}[htb!]
    \centering
    \begin{subfigure}{\textwidth}
        \includegraphics[width=\linewidth]{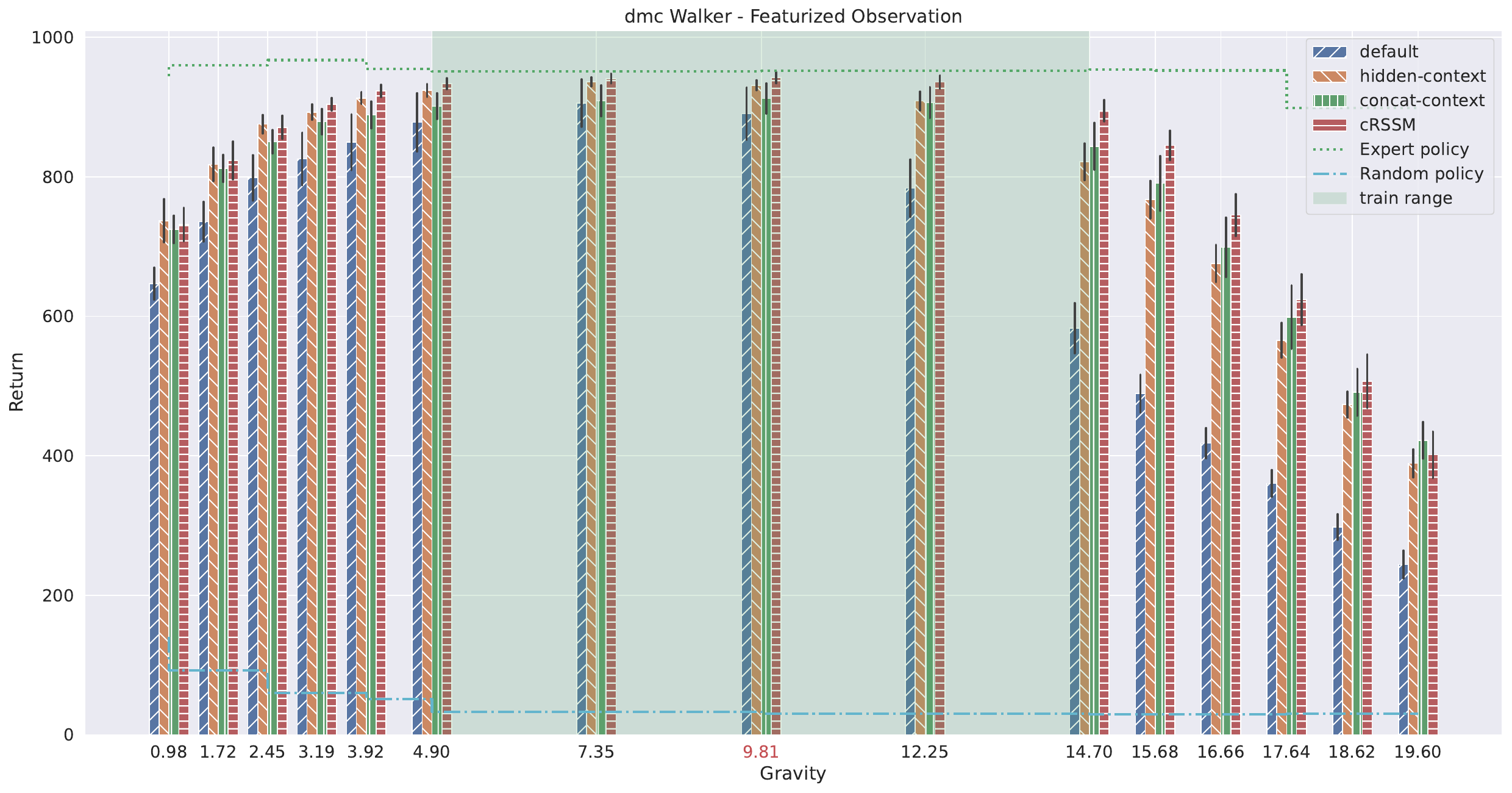}
    \end{subfigure}
    \medskip

    \begin{subfigure}{\textwidth}
        \includegraphics[width=\linewidth]{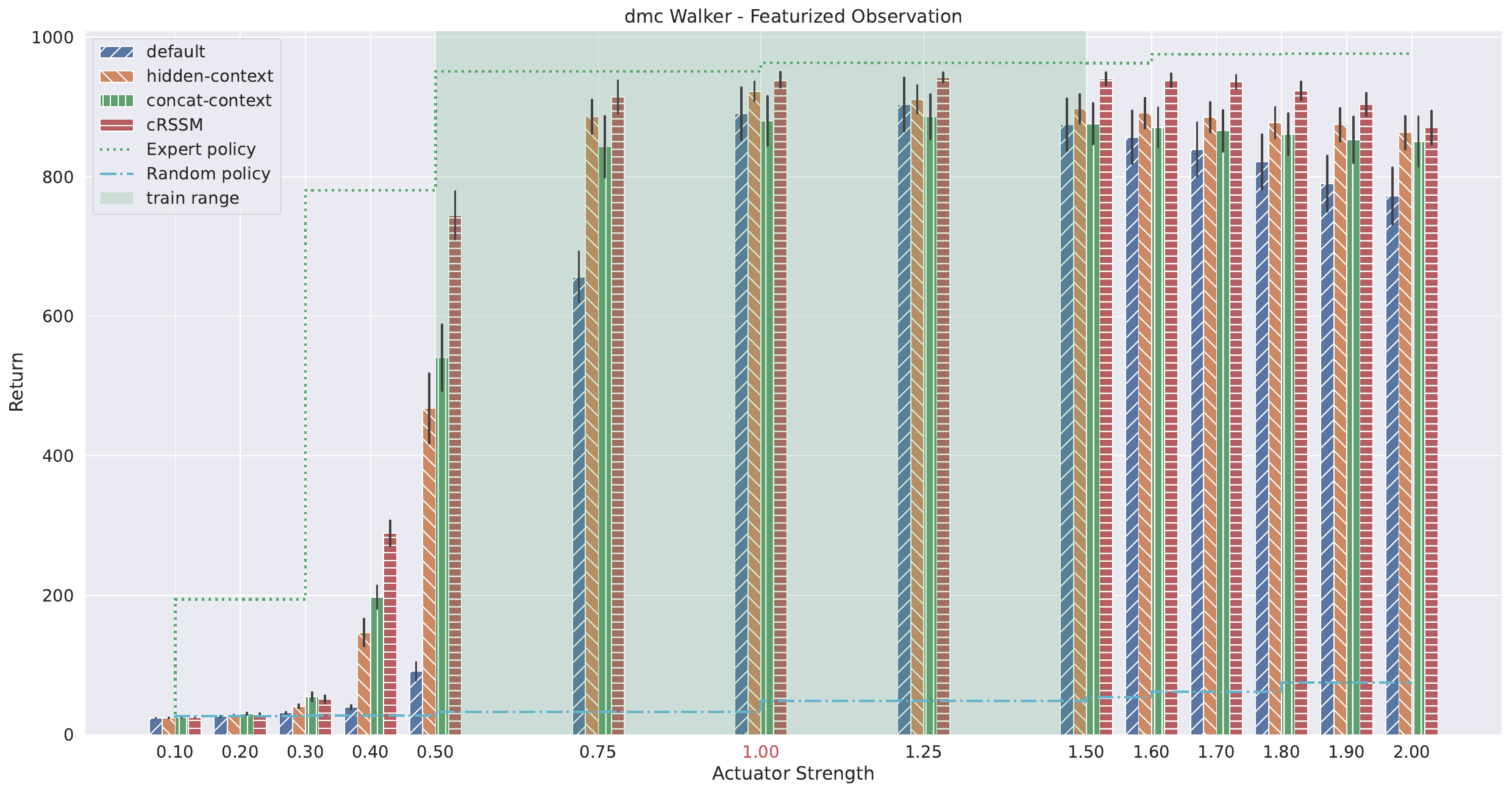}
    \end{subfigure}

    \medskip
    \caption{DMC Walker - Featurized Observations - The mean and standard error of the average evaluation returns are computed across 10 seeds, for 50 evaluation episodes each}\label{fig:walker-feat-appendix}
\end{figure}

\begin{figure}[htb!]
    \centering
    
    \begin{subfigure}{\textwidth}
        \includegraphics[width=\linewidth]{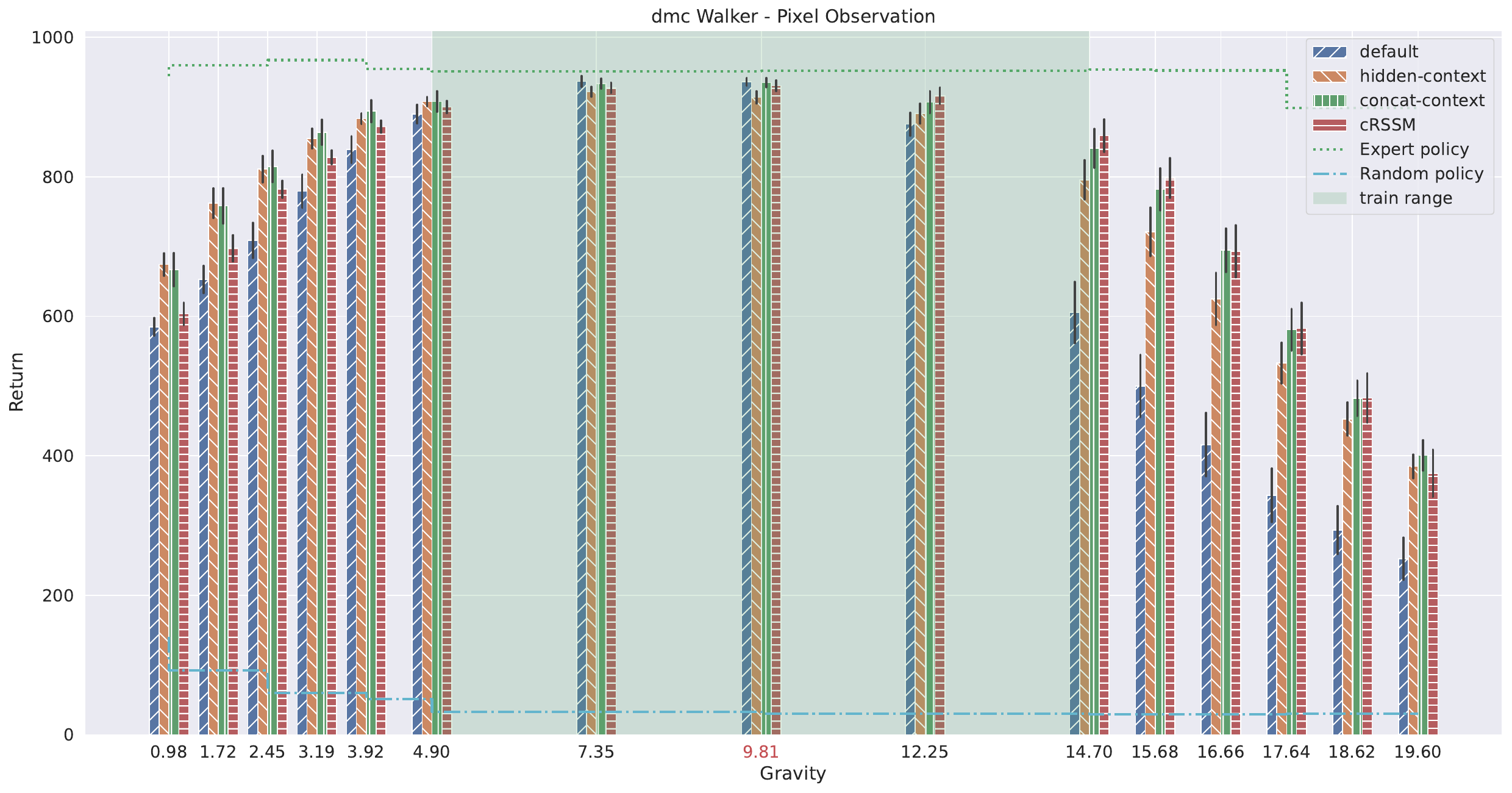}
    \end{subfigure}
    \begin{subfigure}{\textwidth}
        \includegraphics[width=\linewidth]{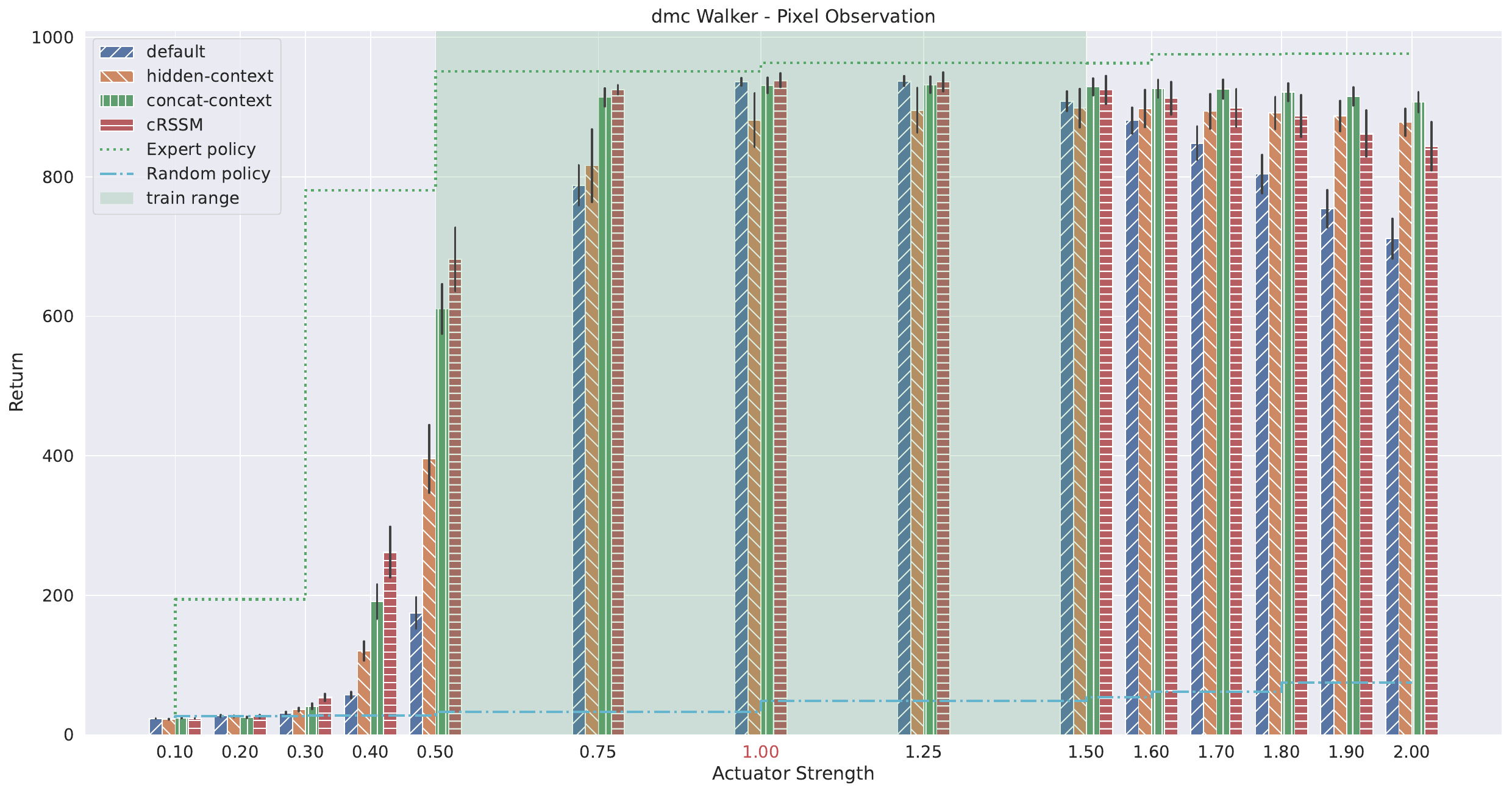}
    \end{subfigure}

    \medskip
    \caption{DMC Walker - Pixel Observations - The mean and standard error of the average evaluation returns are computed across 10 seeds, for 50 evaluation episodes each}
\end{figure}

\clearpage

\subsection{Varying two contexts}

\begin{figure}[htb!]
    \centering
    \begin{subfigure}{\textwidth}
        \includegraphics[width=0.49\linewidth]{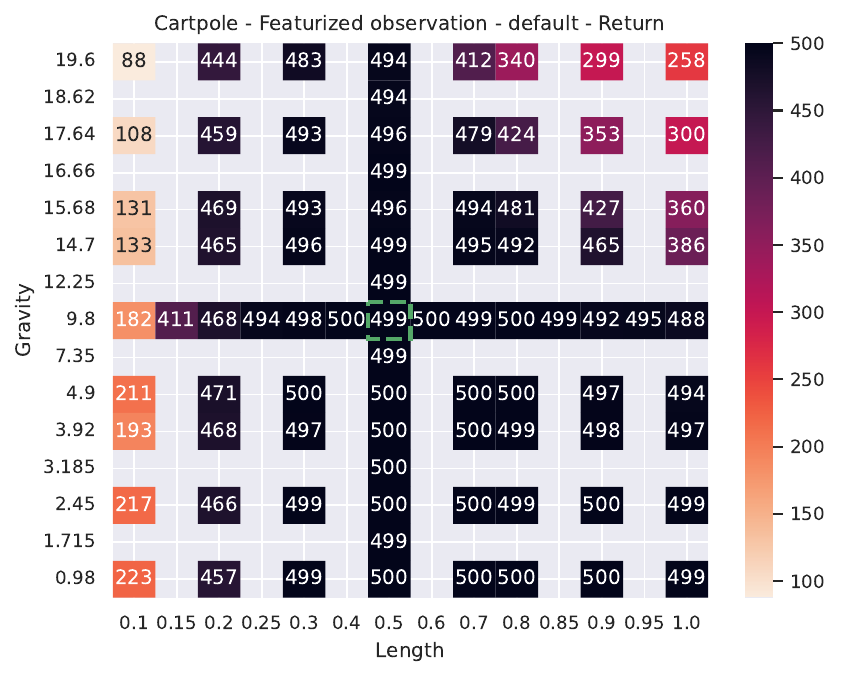}
        \includegraphics[width=0.49\linewidth]{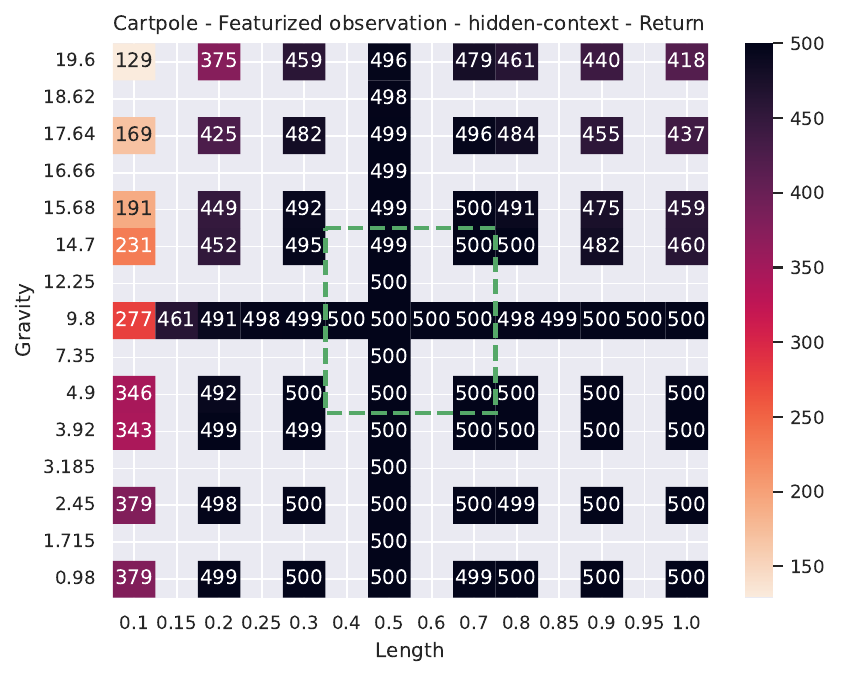}
        \medskip
        \includegraphics[width=0.49\linewidth]{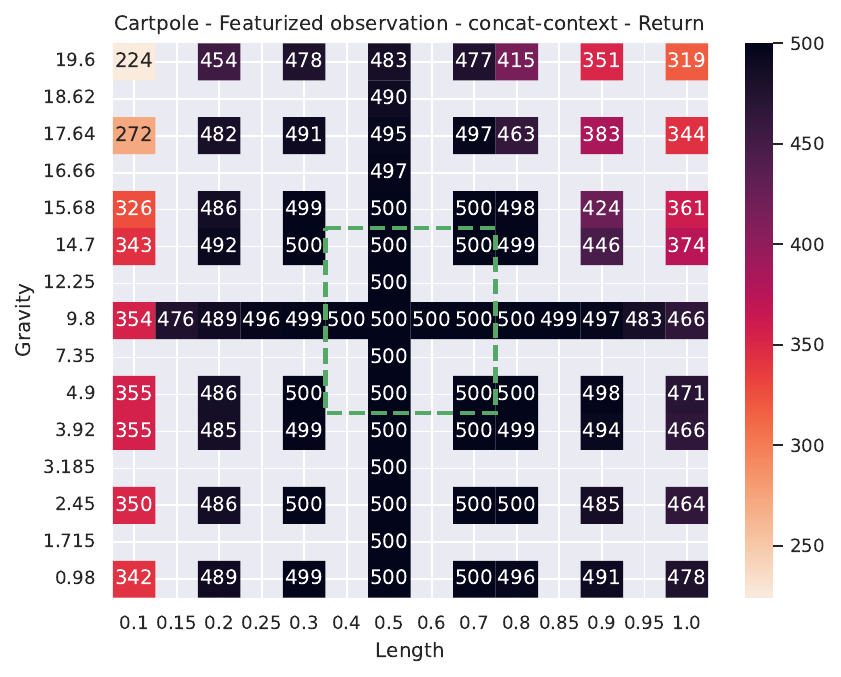}
        \includegraphics[width=0.49\linewidth]{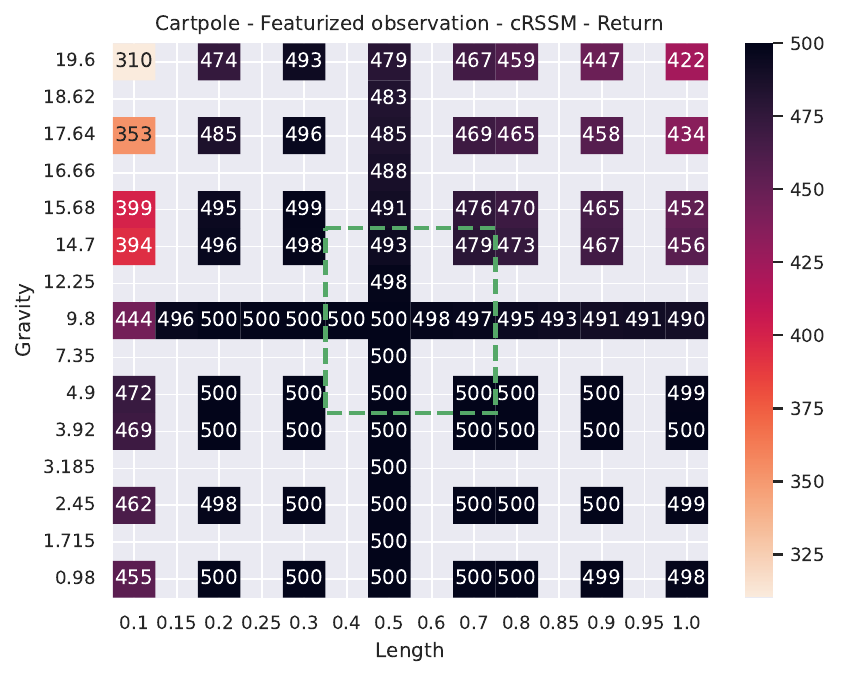}
    \end{subfigure}
    \caption{CartPole - Featurized - The mean and standard error of the average evaluation returns are computed across 10 seeds, for 50 evaluation episodes each}
\end{figure}
\begin{figure}[htb!]
    \begin{subfigure}{\textwidth}
        \includegraphics[width=0.49\linewidth]{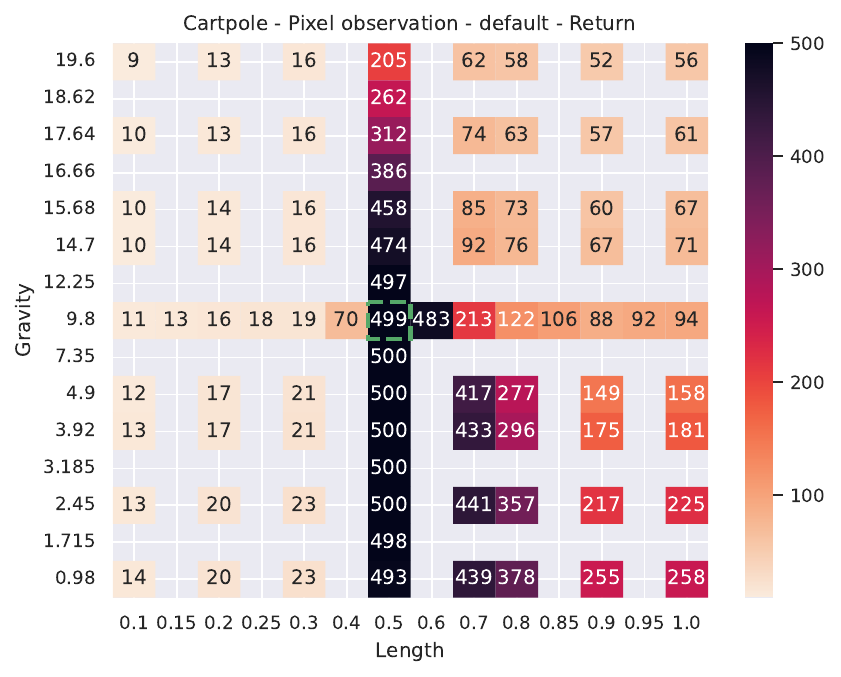}
        \includegraphics[width=0.49\linewidth]{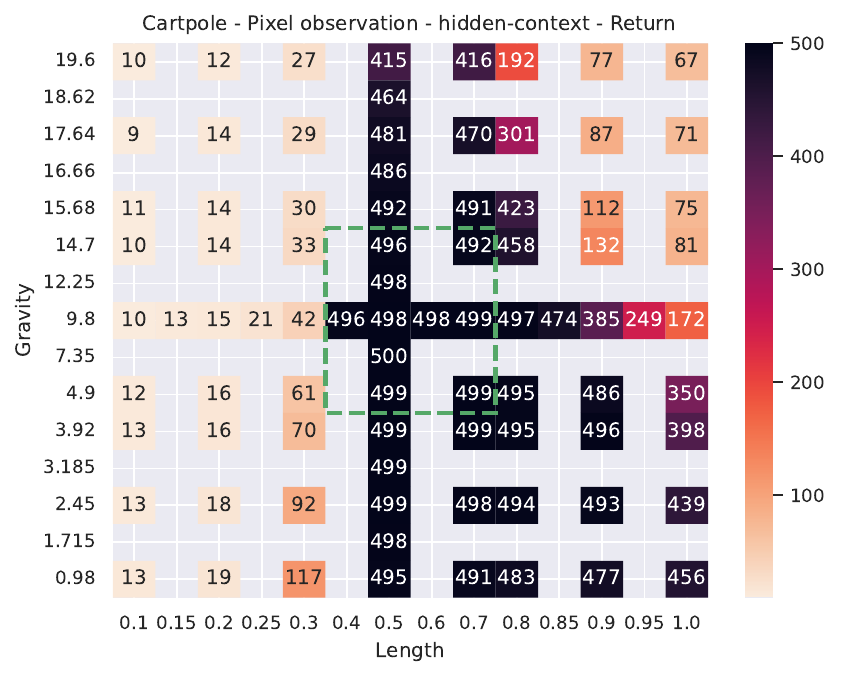}
        \medskip
        \includegraphics[width=0.49\linewidth]{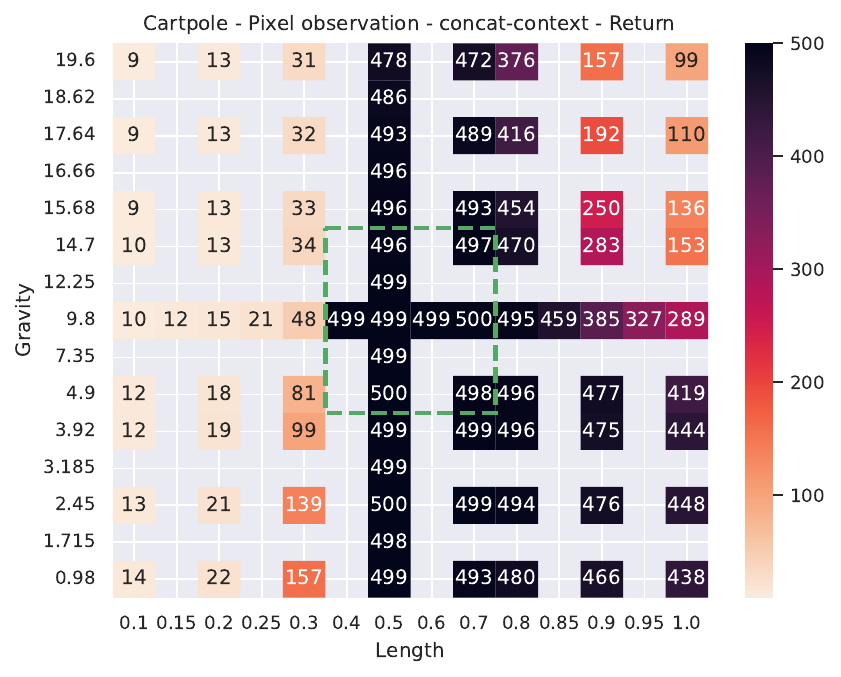}
        \includegraphics[width=0.49\linewidth]{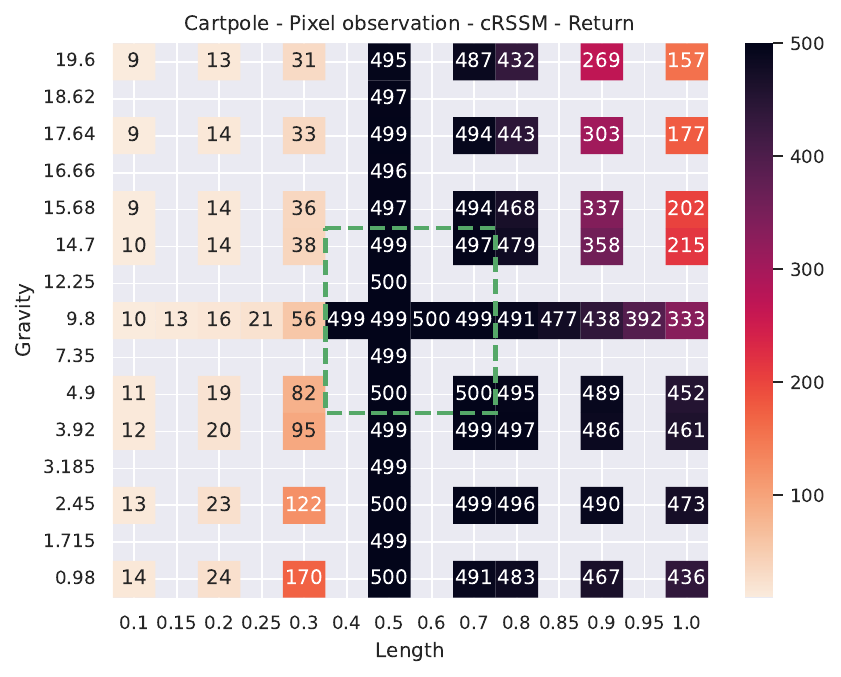}
    \end{subfigure}
    \caption{CartPole - Pixel - The mean and standard error of the average evaluation returns are computed across 10 seeds, for 50 evaluation episodes each}
\end{figure}
\begin{figure}[htb!]
    \centering
    \begin{subfigure}{\textwidth}
        \includegraphics[width=0.49\linewidth]{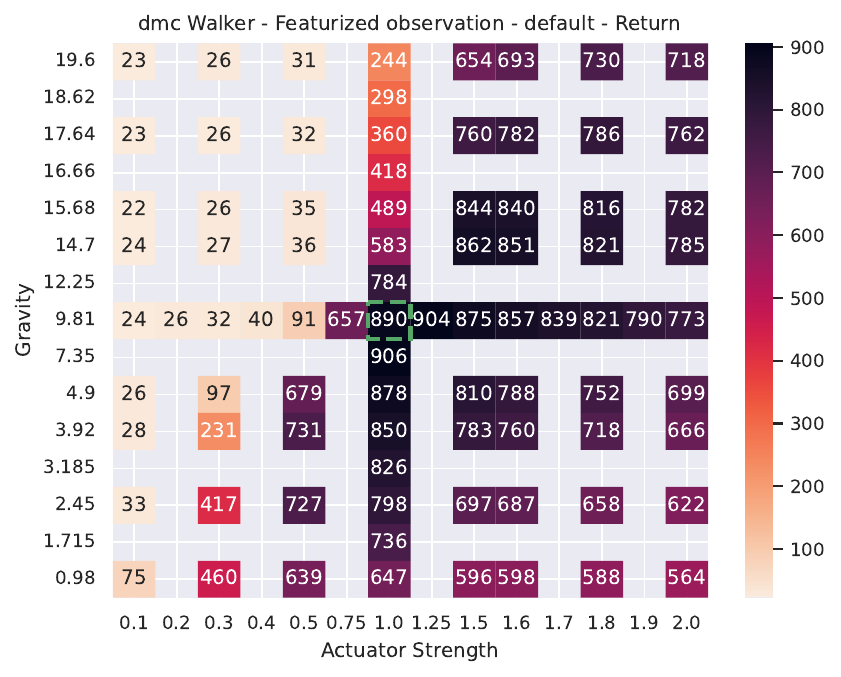}
        \includegraphics[width=0.49\linewidth]{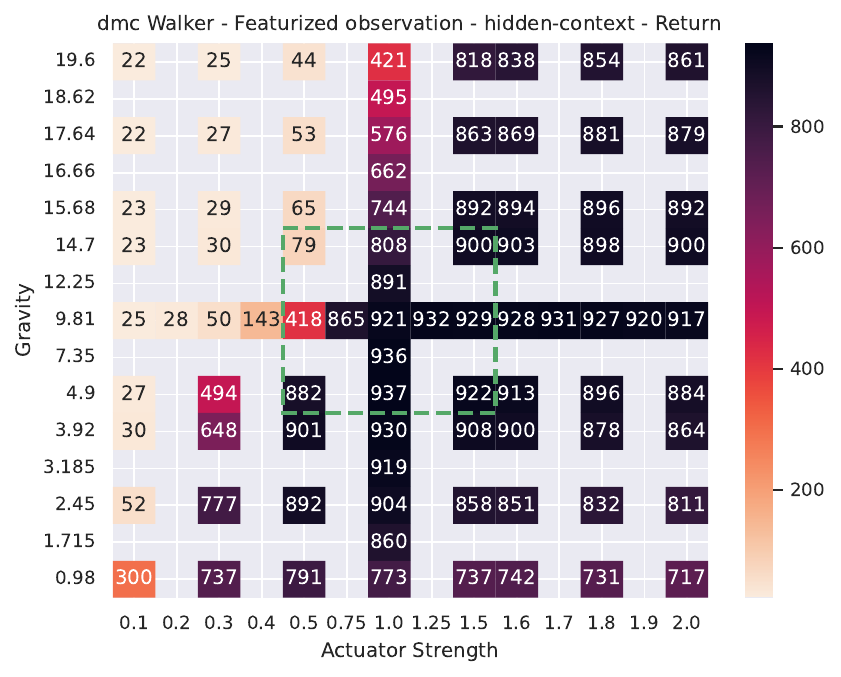}
        \medskip
        \includegraphics[width=0.49\linewidth]{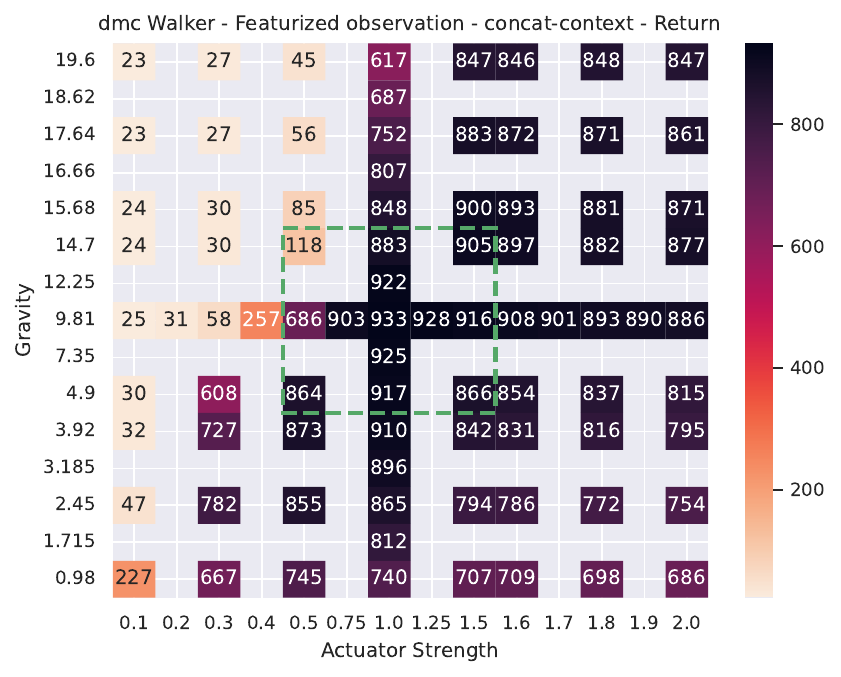}
        \includegraphics[width=0.49\linewidth]{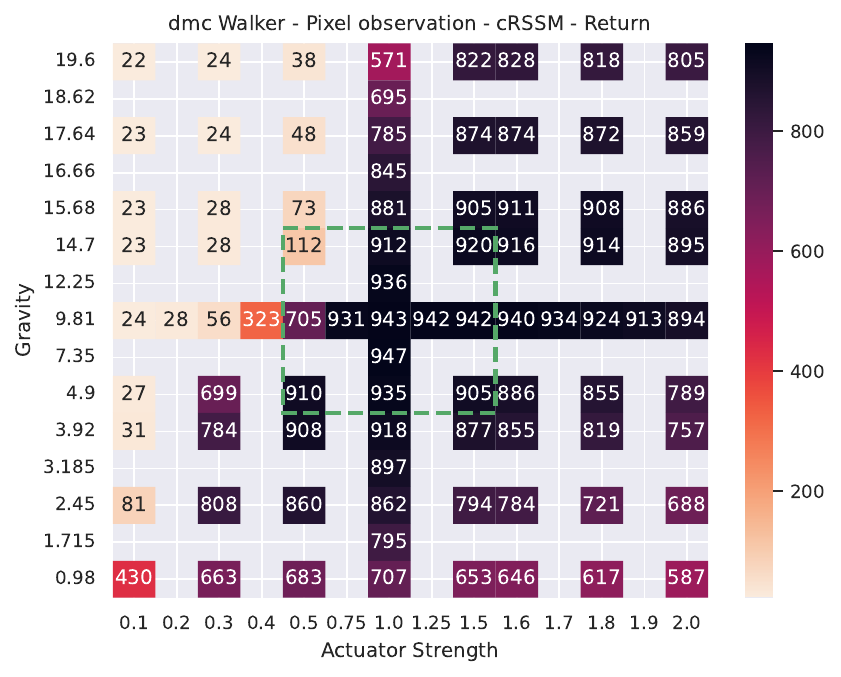}
    \end{subfigure}
    \caption{DMC Walker - Featurized - The mean and standard error of the average evaluation returns are computed across 10 seeds, for 50 evaluation episodes each}
\end{figure}
\begin{figure}
    \centering
    \begin{subfigure}{\textwidth}
        \includegraphics[width=0.49\linewidth]{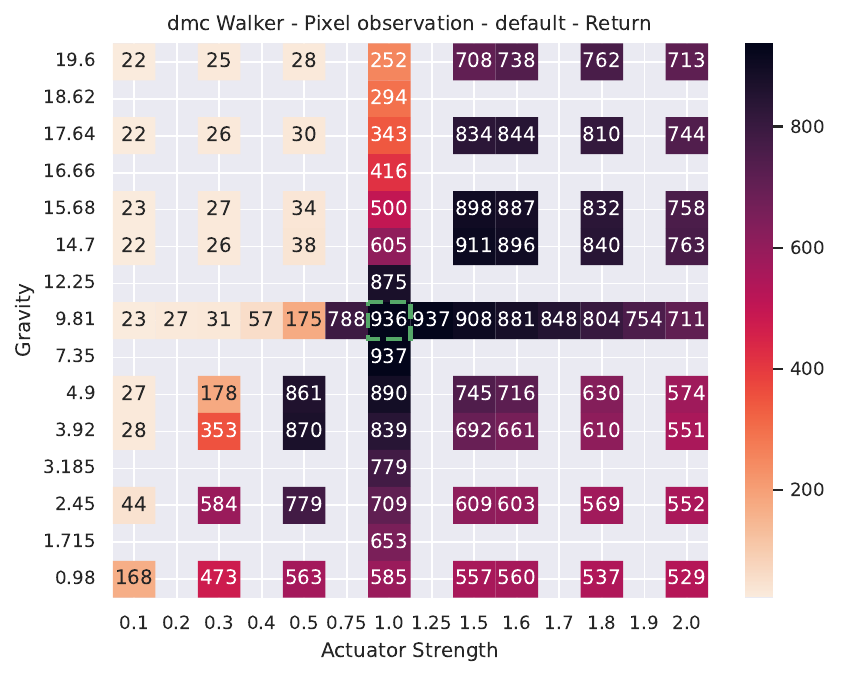}
        \includegraphics[width=0.49\linewidth]{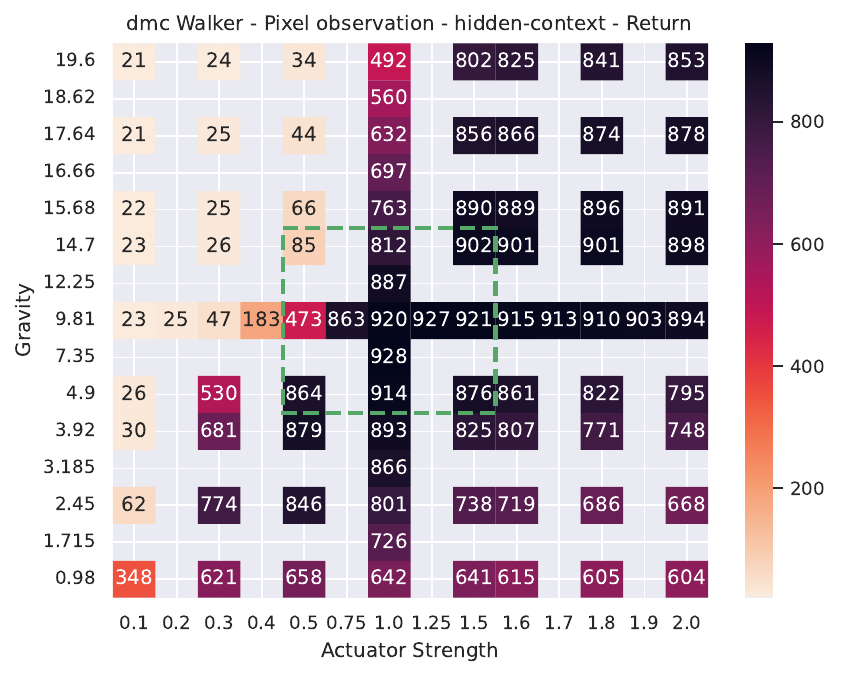}
        \medskip
        \includegraphics[width=0.49\linewidth]{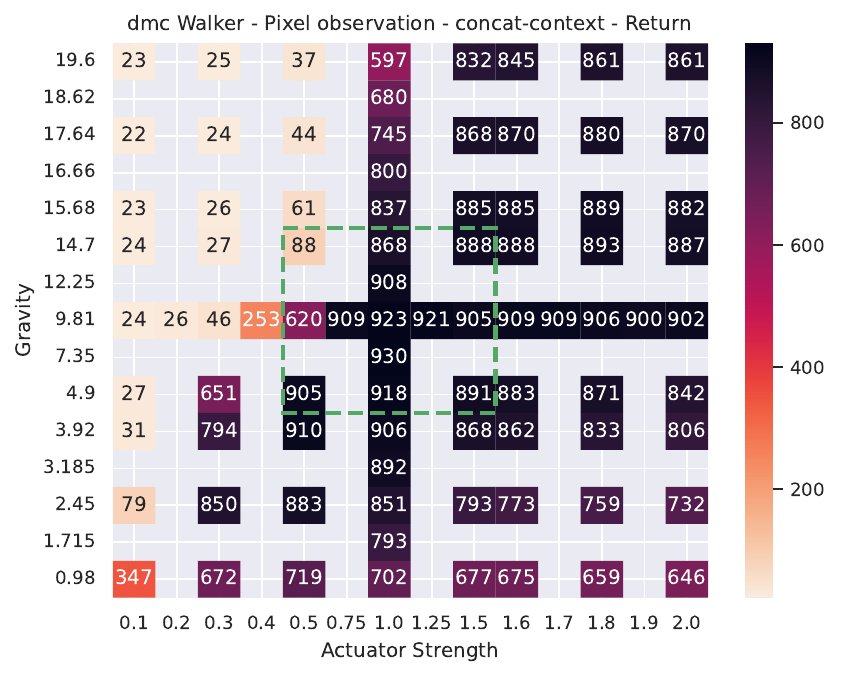}
        \includegraphics[width=0.49\linewidth]{figures/return_plots/dmc_walker_double_box_pixel_observation_pgm_ctx_return.pdf}
    \end{subfigure}
    \caption{DMC Walker - Pixel - The mean and standard error of the average evaluation returns are computed across 10 seeds, for 50 evaluation episodes each}
\end{figure}
\clearpage
\subsection{Probability of Improvement for \crssm}\label{subsec:POI}

In assessing the robustness of an algorithm's improvement over another, considering the average probability of improvement emerges as a valuable metric. Specifically, it calculates the probability of Algorithm X surpassing Algorithm Y on a randomly chosen task, disregarding the magnitude of improvement. Identifying the optimal aggregate metric remains an ongoing inquiry, and presenting multiple metrics, which circumvent the pitfalls of prevalent metrics, ensures reliability and efficiency in decision-making processes.
\begin{figure}[h]
    \centering
    \includegraphics[width=0.98\linewidth]{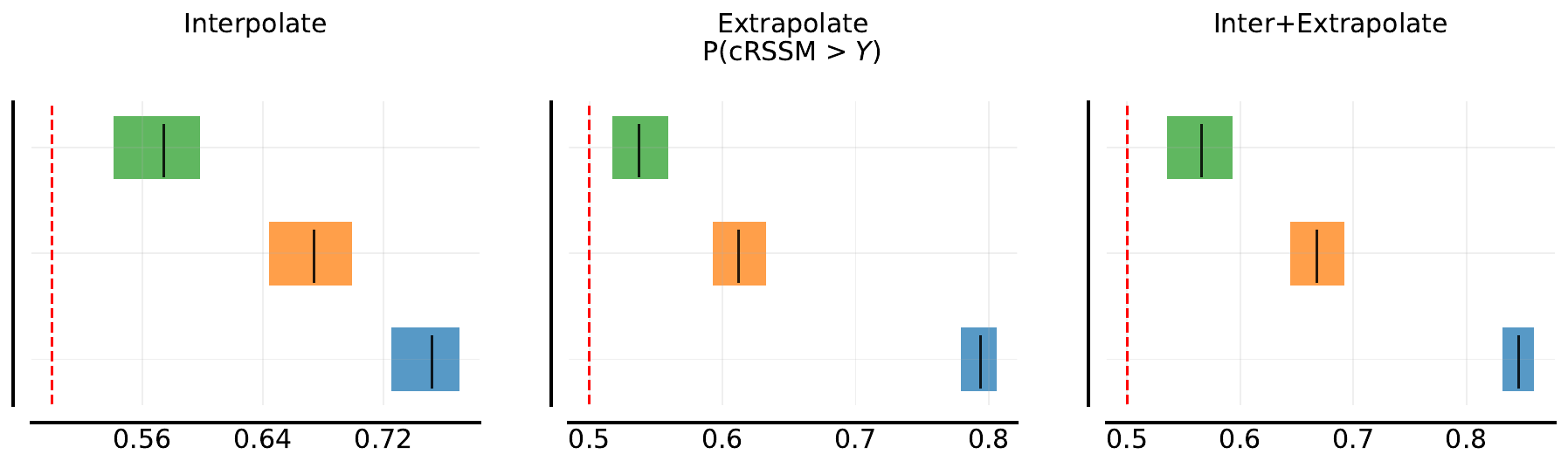}
    \caption{Aggregate probability of improvement for pixel modality. }
    \label{fig:POI_p}
\end{figure}
\begin{figure}[h]
    \centering
    \includegraphics[width=0.98\linewidth]{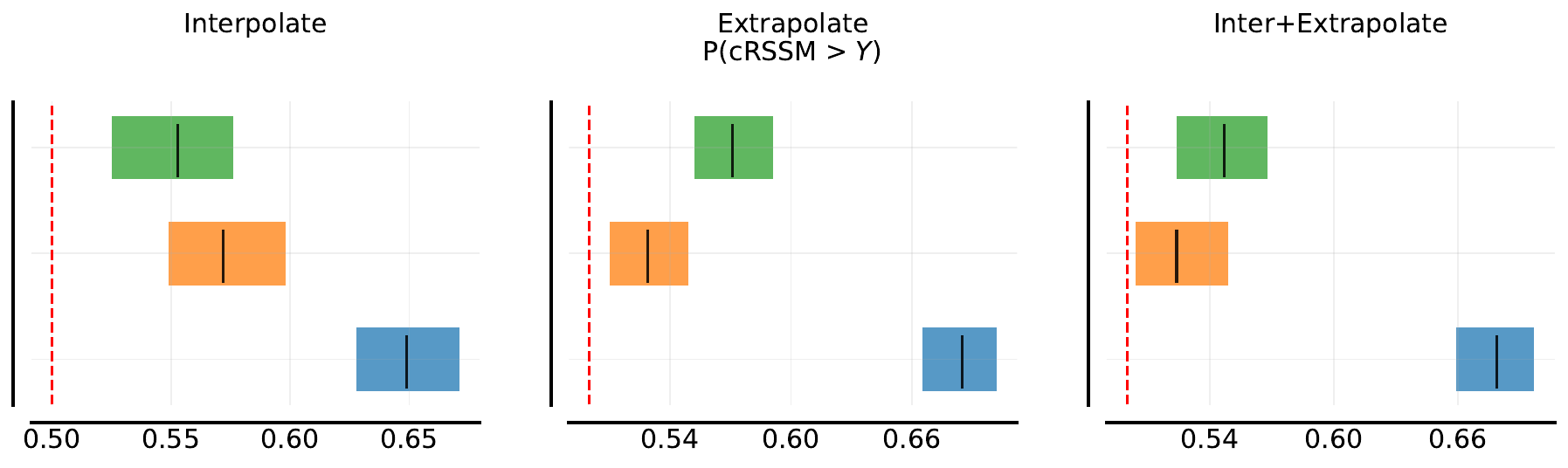}
    \caption{Aggregate probability of improvement for the featurized modality.}
    \label{fig:POI_f}
\end{figure}


\subsection{Expert Normalized IQM Plots for Individual Settings}
\label{sec:individual_icm}
The IQM plots corresponding to the settings in \ref{tab:quantitative_results}. For some settings in the Cartpole environment, since we reach optimal expert performance across all seeds, the plots look empty.

\begin{figure}[tbp]
    \centering
    \begin{subfigure}[b]{\textwidth} 
        \includegraphics[width=\textwidth]{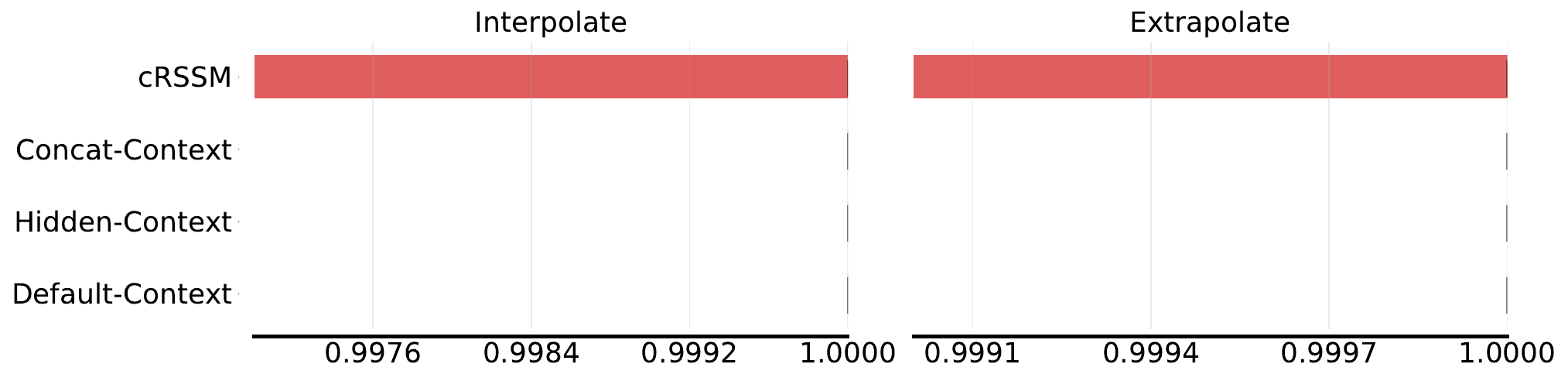}%
    \end{subfigure}

    \caption{Cartpole - Featurized - Gravity: Expert normalized IQM with 95\% confidence interval}
\end{figure}

\begin{figure}[tbp]
    \centering
    \begin{subfigure}[b]{\textwidth} 
        \includegraphics[width=\textwidth]{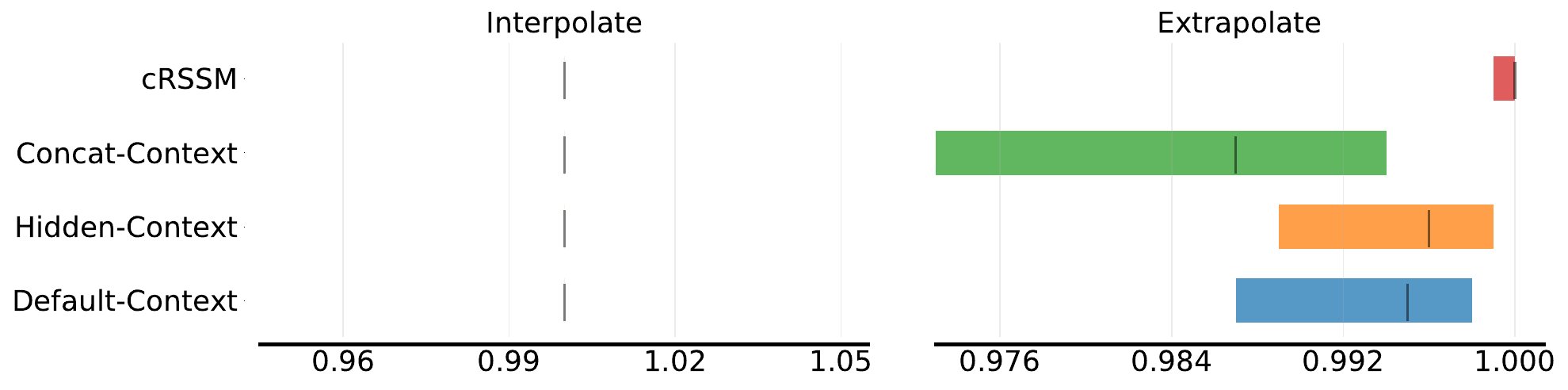}%
    \end{subfigure}

    \caption{Cartpole - Featurized - Length: Expert normalized IQM with 95\% confidence interval}
\end{figure}

\begin{figure}[tbp]
    \centering
    \begin{subfigure}[b]{\textwidth} 
        \includegraphics[width=\textwidth]{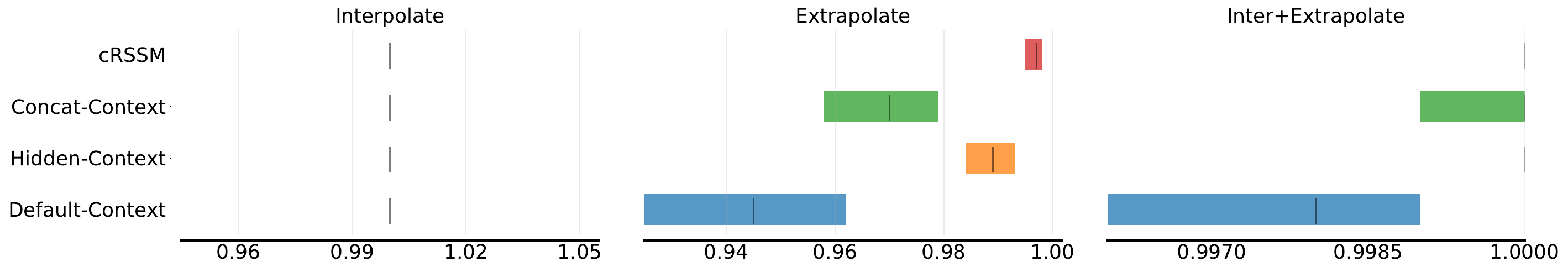}%
    \end{subfigure}

    \caption{Cartpole - Featurized -  Gravity + Length: Expert normalized IQM with 95\% confidence interval}
\end{figure}

\begin{figure}[tbp]
    \centering
    \begin{subfigure}[b]{\textwidth} 
        \includegraphics[width=\textwidth]{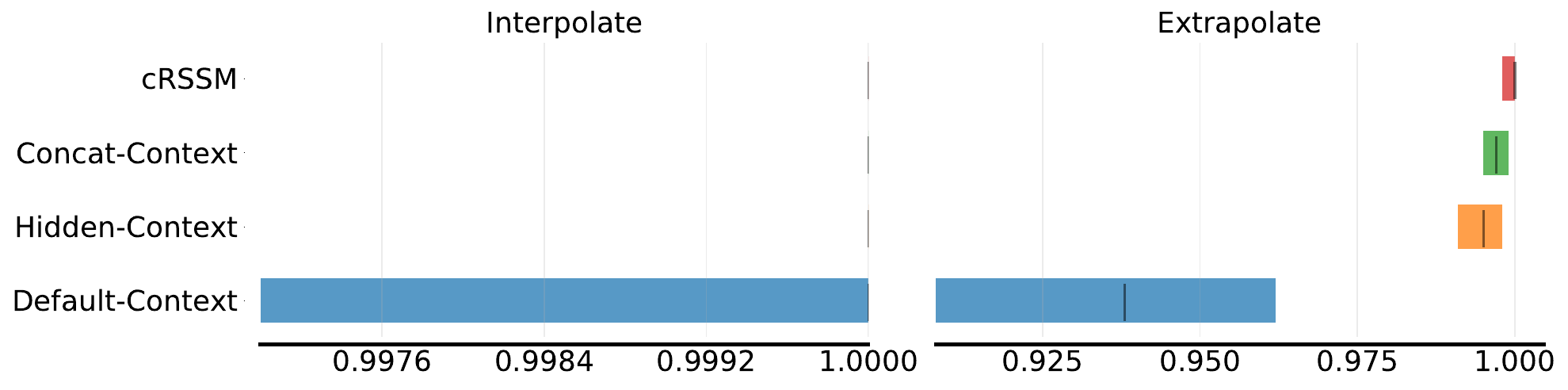}%
    \end{subfigure}

    \caption{Cartpole - Pixel - Gravity: Expert normalized IQM with 95\% confidence interval}
\end{figure}

\begin{figure}[tbp]
    \centering
    \begin{subfigure}[b]{\textwidth} 
        \includegraphics[width=\textwidth]{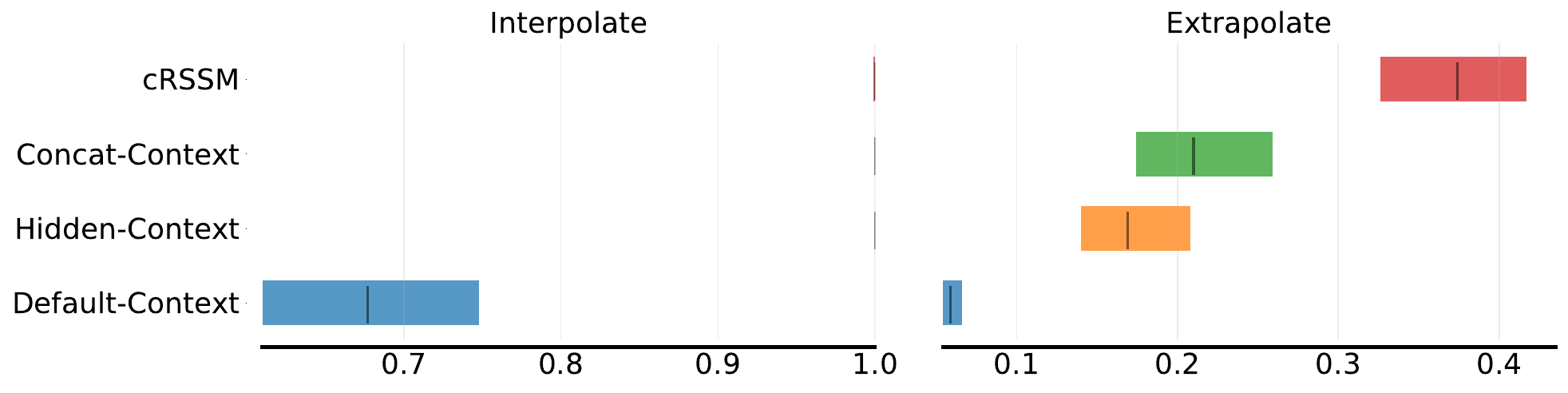}%
    \end{subfigure}

    \caption{Cartpole - Pixel - Length: Expert normalized IQM with 95\% confidence interval}
\end{figure}

\begin{figure}[tbp]
    \centering
    \begin{subfigure}[b]{\textwidth} 
        \includegraphics[width=\textwidth]{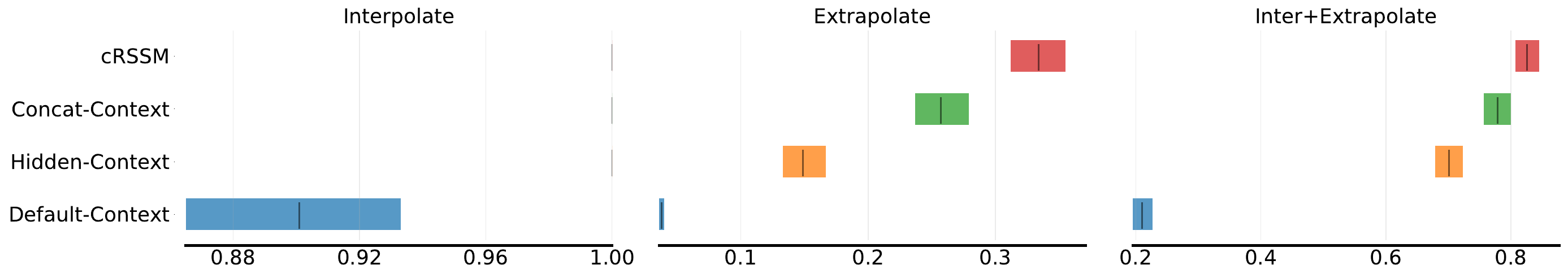}%
    \end{subfigure}

    \caption{Cartpole - Pixel - Gravity + Length: Expert normalized IQM with 95\% confidence interval}
\end{figure}

\begin{figure}[tbp]
    \centering
    \begin{subfigure}[b]{\textwidth} 
        \includegraphics[width=\textwidth]{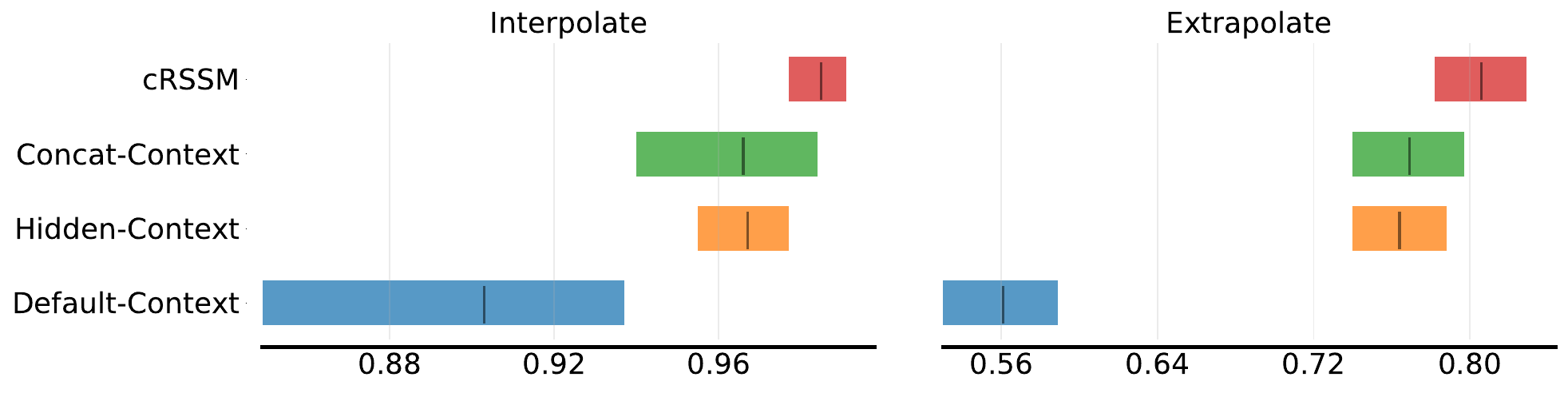}%
    \end{subfigure}

    \caption{DMC Walker - Featurized - Gravity: Expert normalized IQM with 95\% confidence interval}
\end{figure}

\begin{figure}[tbp]
    \centering
    \begin{subfigure}[b]{\textwidth} 
        \includegraphics[width=\textwidth]{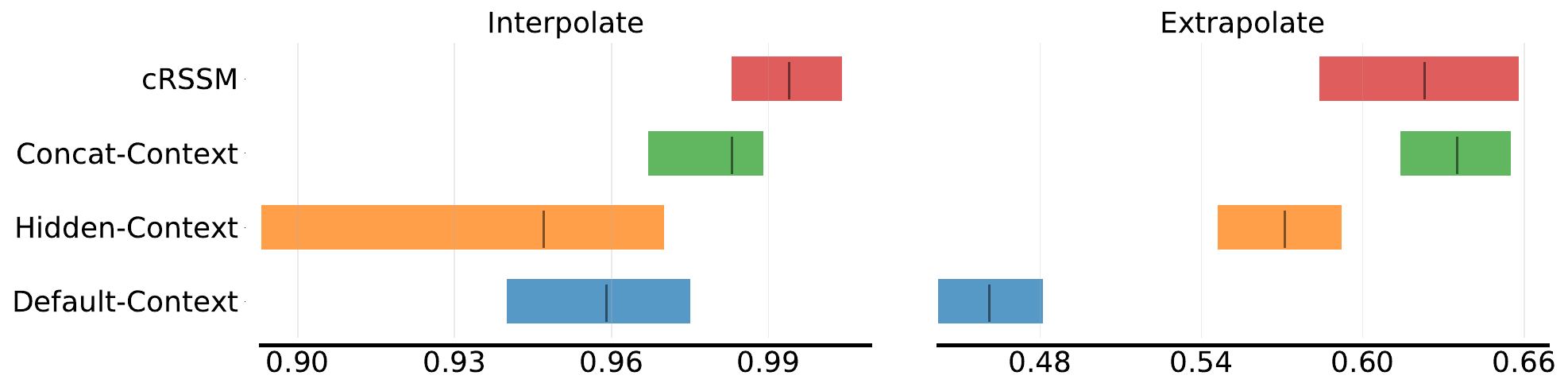}%
    \end{subfigure}

    \caption{DMC Walker - Featurized - Actuator Strength: Expert normalized IQM with 95\% confidence interval}
\end{figure}

\begin{figure}[tbp]
    \centering
    \begin{subfigure}[b]{\textwidth} 
        \includegraphics[width=\textwidth]{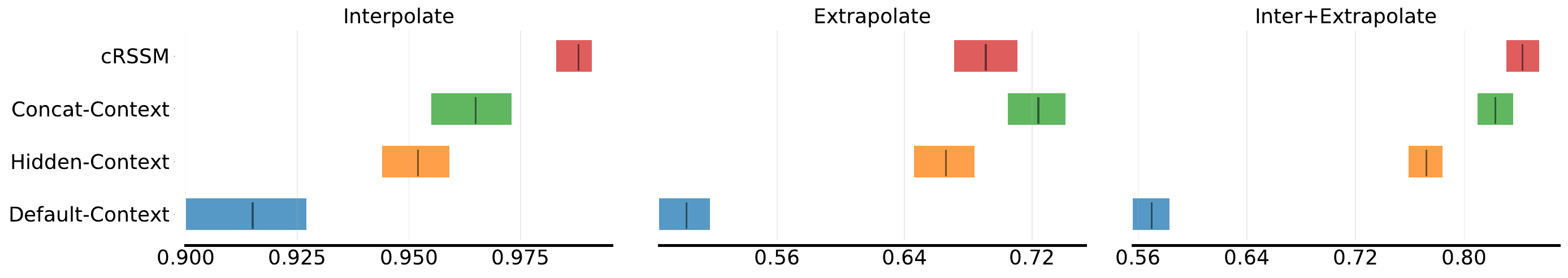}%
    \end{subfigure}

    \caption{DMC Walker - Featurized - Gravity + Actuator Strength: Expert normalized IQM with 95\% confidence interval}
\end{figure}

\begin{figure}[tbp]
    \centering
    \begin{subfigure}[b]{\textwidth} 
        \includegraphics[width=\textwidth]{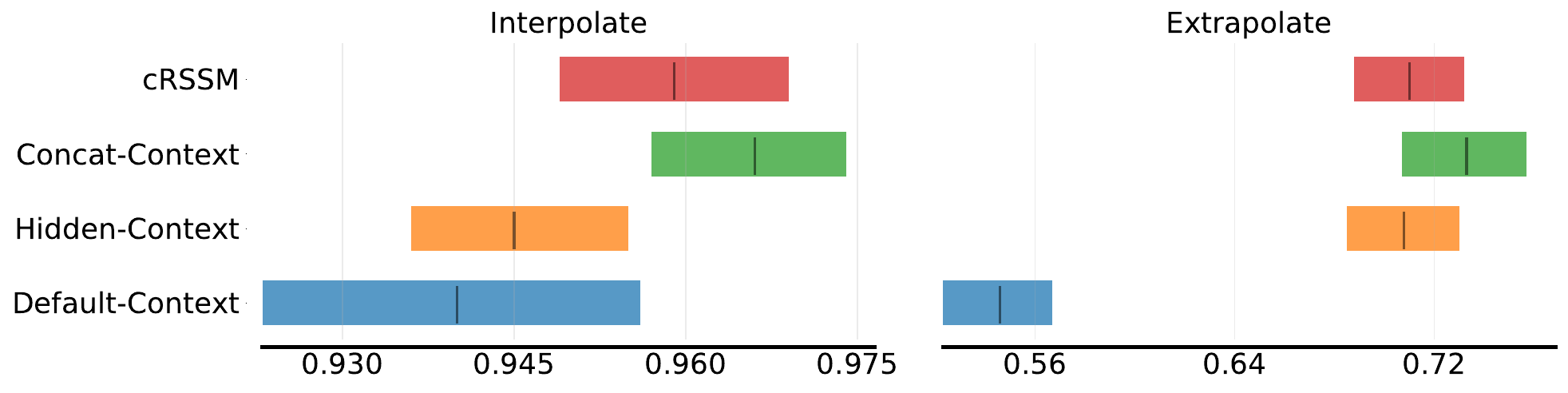}%
    \end{subfigure}

    \caption{DMC Walker - Pixel - Gravity: Expert normalized IQM with 95\% confidence interval}
\end{figure}

\begin{figure}[tbp]
    \centering
    \begin{subfigure}[b]{\textwidth} 
        \includegraphics[width=\textwidth]{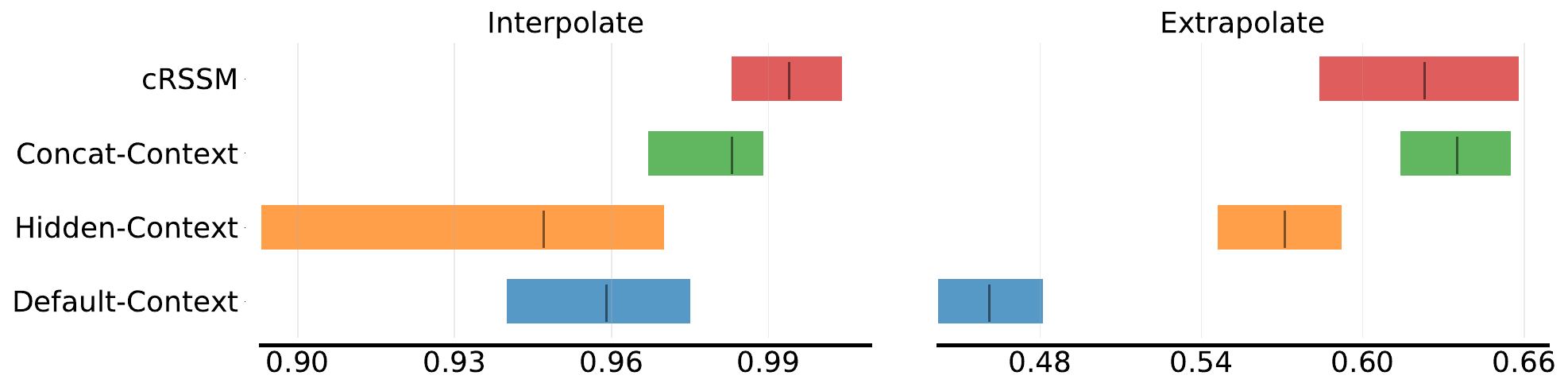}%
    \end{subfigure}

    \caption{DMC Walker - Pixel - Actuator Strength: Expert normalized IQM with 95\% confidence interval}
\end{figure}

\begin{figure}[tbp]
    \centering
    \begin{subfigure}[b]{\textwidth} 
        \includegraphics[width=\textwidth]{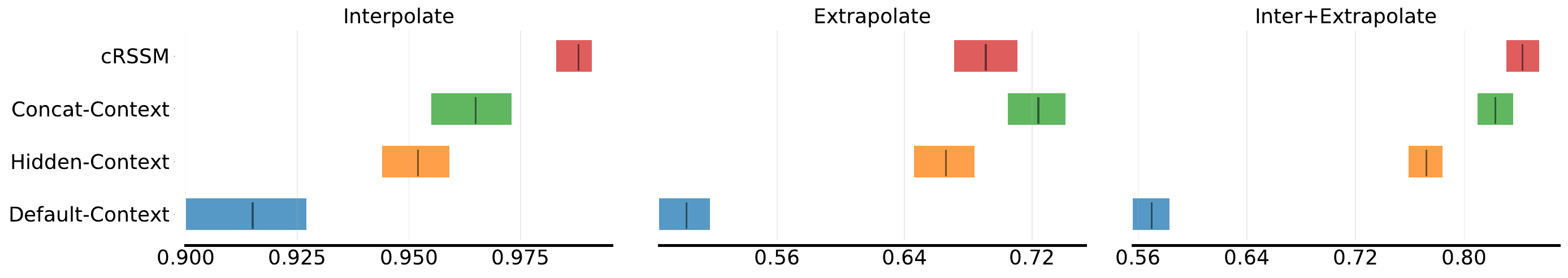}%
    \end{subfigure}

    \caption{DMC Walker - Pixel - Gravity + Actuator Strength: Expert normalized IQM with 95\% confidence interval}
\end{figure}

\clearpage
\section{Intuitive Interpretation of the RSSM \& cRSSM}
\label{sec:intutive_crssm}

We can use a video game analogy to gain an intuitive understanding of the RSSM and cRSSM. The context is the game settings, such as difficulty, which do not change while playing the game. The deterministic state is the memory of the game engine. The stochastic state models the aleatoric uncertainty used by the game engine, i.e. whenever the game samples from a random number generator and certain variables of the current game state, it is the same as sampling from the stochastic state model. After this sampling is done and the user inputs the action, the deterministic state model is akin to the game's logic, which uses the context of the current game memory state and the sampled stochastic state to compute the next game memory state. The observation model is the game engine's visual renderer that maps the game memory to the pixels you see on your monitor. It can also use context to render things differently. Finally, the reward model is the score, a distribution conditioned on the state and context, as the context (say, difficulty setting) can influence how many points you get for a given state.
\clearpage
\section{Hyperparameters}
\label{app:hparam}
We choose the \textit{small} variant of DreamerV3 with all hyperparameters taken from \cite{hafner-arxiv23a}.

\begin{table}[htb!]
\centering
\begin{tabular}{lr}
\toprule
\textbf{Name} & \textbf{Value} \\
\midrule
\multicolumn{2}{l}{\textbf{General}} \\
\midrule
Replay capacity (FIFO) & $10^6\!\!$ \\
Batch size &  16 \\
Batch length & 64 \\
Activation & $\operatorname{LayerNorm}+\operatorname{SiLU}$ \\
\midrule
\multicolumn{2}{l}{\textbf{World Model}} \\
\midrule
Deterministic State model (GRU) units & 512 \\
MLP layers & 2 \\
MLP units & 512 \\
Number of latents  & 32 \\
Classes per latent & 32 \\
Reconstruction loss scale & 1.0 \\
Dynamics loss scale  & 0.5 \\
Representation loss scale & 0.1 \\
Learning rate & $10^{-4}$ \\
Adam epsilon  & $10^{-8}$ \\
Gradient clipping & 1000 \\
\midrule
\multicolumn{2}{l}{\textbf{Actor Critic}} \\
\midrule
MLP layers & 2 \\
MLP units & 512 \\
Imagination horizon & 15 \\
Discount horizon & 333 \\
Return lambda & 0.95 \\
Critic EMA decay  & 0.98 \\
Critic EMA regularizer & 1 \\
Return normalization scale  & $\operatorname{Per}(R, 95) - \operatorname{Per}(R, 5)$ \\
Return normalization limit & 1 \\
Return normalization decay & 0.99 \\
Actor entropy scale & $3\cdot10^{-4}$ \\
Learning rate & $3\cdot10^{-5}$ \\
Adam epsilon  & $10^{-5}$ \\
Gradient clipping & 100 \\
\bottomrule

\end{tabular}
\caption{DreamerV3 hyper parameters. The same values are used across all experiments.
}
\label{tab:hparams}
\end{table}

\clearpage
\section{Results for DMC Walker - 100k Steps}\label{sec:walker_less_samples}

For comprehensive evaluations, we conducted intermediate assessments on the DMC walker environment, using 10 seeds for 100k environment steps in each generalization setting. Normalized IQM scores, detailed in \Cref{tab:walker_100k}, demonstrate superior performance in the most challenging featurized cases with both contexts. However, within some settings, particularly for the pixel modality, we observed a notable lag in performance. This discrepancy, especially in the interpolation region, where the evaluation distribution aligns closely with the training distribution, indicates the need for additional samples to facilitate effective learning. We present the complete training (500k steps) in Table \ref{tab:quantitative_results}.

\begin{table}[!htb]
    \centering
    \begin{tabular}{lccc|ccc}
        \toprule
                \multicolumn{7}{c}{Walker - 100k steps}  \\ 
        \cmidrule(lr){2-7}
                    &    \multicolumn{3}{c}{Featurized} &  \multicolumn{3}{c}{Pixel}  \\ 
        \cmidrule(lr){2-7}
        (g {\color{default}d}) & 0.737 & 0.551 & - & 0.549 & 0.376 & - \\
        (g {\color{hidden}h}) & 0.682 & 0.547 & - & \textbf{0.697} & 0.489 & - \\
        (g {\color{concat}c}) & \textbf{0.864} & \textbf{0.684} & - & 0.656 & \textbf{0.520} & - \\
        (g {\color{crssm}cR}) & 0.779 & 0.661 & - & 0.565 & 0.450 & - \\
         \midrule
        (a {\color{default}d}) & 0.824 & 0.406 & - & 0.634 & 0.305 & - \\
        (a {\color{hidden}h}) & 0.833 & 0.437 & - & 0.674 & 0.376 & - \\
        (a {\color{concat}c}) & 0.741 & 0.402 &- & 0.732 & \textbf{0.409} & - \\
        (a {\color{crssm}cR}) & \textbf{0.947} & \textbf{0.456} & - & \textbf{0.819} & 0.391 & - \\
        \midrule
        (g+a {\color{default}d}) & 0.749 & 0.411 & 0.537 & 0.574 & 0.326 & 0.372 \\
        (g+a {\color{hidden}h}) & 0.722 & 0.401 & 0.521 & 0.649 & 0.396 & 0.469 \\
        (g+a {\color{concat}c}) & 0.652 & 0.357 & 0.470 & \textbf{0.658} & \textbf{0.406} & \textbf{0.494} \\
        (g+a {\color{crssm}cR}) & \textbf{0.884} & \textbf{0.447} & \textbf{0.632} & 0.606 & 0.333 & 0.423 \\ \bottomrule
    \end{tabular}%
        \caption{Expert normalized IQM over 10 seeds for different evaluation settings, in featurized and pixel modality. Each described by three variables: context, method, and mode. Context takes values from $\{ g: \text{{gravity}}, a: \text{{actuator strength}}, l: \text{{pole length}} \}$ with + indicating multiple contexts; and method from $\{ {\color{default}d: \text{{\defaultagent}}}, {\color{hidden}h: \text{{\hiddenagent}}}, {\color{concat}c: \text{{\concatagent}}}, {\color{crssm}cR: \text{{\crssm}}} \}$}%
        \label{tab:walker_100k}
\end{table}%

\begin{figure}[htb!]
    \begin{subfigure}{0.45\textwidth}
        \centering\includegraphics[width=\linewidth]{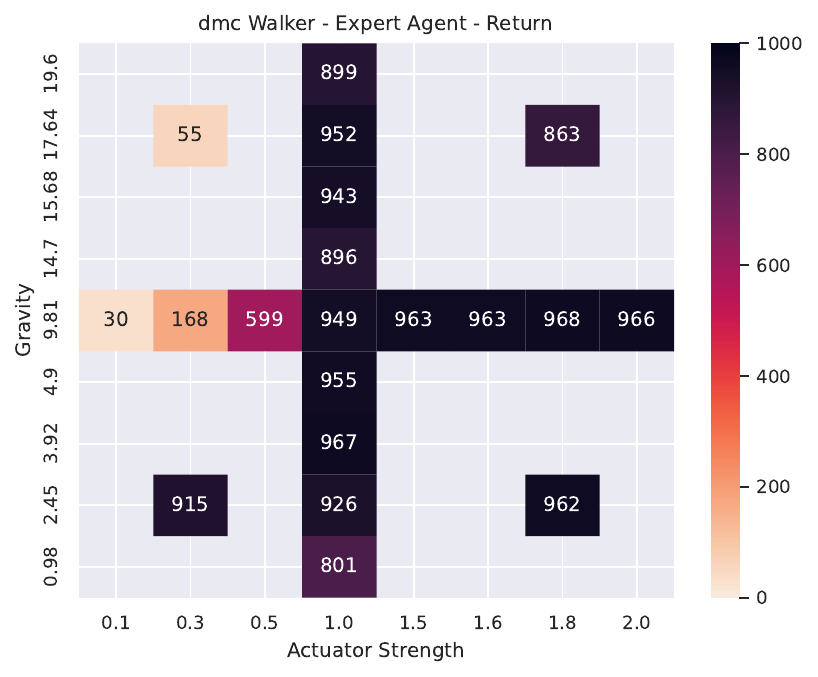}
    \end{subfigure}
    \caption{The best expert trained on each context over 5 seeds for 100k steps on DMC walker. We use featurized modality with less partial observability compared to pixels, to get an optimistic upper bound of expert returns.}
\end{figure}

\section{Discussion on More Comprehensive Benchmarks}
\label{app:benchmarks}
Creating contextual benchmarks for environments such as Atari, DMLab, ProcGen, and Minecraft presents an important opportunity for further research into ZSG. Unlike our current tasks focusing on motor control in environments like Walker and CartPole, some of these benchmarks emphasize different aspects such as navigation and exploration while others such as procedurally generated or open-ended worlds offer dynamic objectives and high variability, requiring strategic planning and adaptability to diverse challenges. These complexities necessitate pronounced policy adjustments with changing contexts. However, creating variants in such benchmarks is challenging due to their inherent intricacies and lack of easily accessible interfaces. By addressing these complexities, future work could unlock critical insights into the adaptability of reinforcement learning agents in diverse and changing conditions.

\end{document}